\documentclass[lettersize,journal]{IEEEtran}
\usepackage{amsmath,amsfonts}
\usepackage{algorithmic}
\usepackage{algorithm}
\usepackage{array}
\usepackage[caption=false,font=normalsize,labelfont=sf,textfont=sf]{subfig}
\usepackage{textcomp}
\usepackage{stfloats}
\usepackage{url}
\usepackage{verbatim}
\usepackage{graphicx}
\usepackage{cite}
\newcommand{\name}{SVGS}
\usepackage{overpic}
\usepackage{color}
\usepackage{multirow}
\usepackage{enumerate}
\usepackage{wrapfig}
\usepackage{amsmath}
\usepackage{colortbl}
\usepackage{xcolor}
\usepackage{enumerate}
\usepackage{soul}
\usepackage{booktabs}
\usepackage{lineno}
\usepackage[numbers,sort]{natbib}
\usepackage{amssymb}
\usepackage{stfloats}
\usepackage{svg}

\newcommand{\XR}[1]{{#1}}

\newcommand{\FF}[1]{{#1}}
\definecolor{rk1}{RGB}{255, 153, 153}
\definecolor{rk2}{RGB}{255, 204, 153}
\definecolor{rk3}{RGB}{254, 248, 173}
\definecolor{rk4}{RGB}{255, 255, 205}
\begin{document}

\title{SVGS: Enhancing Gaussian Splatting Using Primitives with
Spatially Varying Colors}

\author{Rui Xu, Wenyue Chen, Jiepeng Wang, Yuan Liu, Peng Wang, Cheng Lin$^\dagger$, Shiqing Xin, Xin Li,\\ Wenping Wang\IEEEmembership{, Fellow,~IEEE}, Taku Komura$^\dagger$
\IEEEcompsocitemizethanks{
\IEEEcompsocthanksitem Rui Xu, Jiepeng Wang, Peng Wang, Taku Komura are with the Department of Computer Science, The University of Hong Kong. 
\IEEEcompsocthanksitem Wenyue Chen is with Shenzhen Graduate School of Peking University.
\IEEEcompsocthanksitem Yuan Liu is with Nanyang Technological University and Hong Kong University of Science and Technology.
\IEEEcompsocthanksitem Cheng Lin is with the Department of Computer Science and Engineering of Macau University of Science and Technology.
\IEEEcompsocthanksitem Shiqing Xin is with the School of Computer Science, Shandong University.
\IEEEcompsocthanksitem Xin Li and Wenping Wang are with Texas A\&M University.
\IEEEcompsocthanksitem Taku Komura$^\dagger$ and Cheng Lin$^\dagger$ are the co-corresponding authors. (Email: taku@cs.hku.hk, chenglin@must.edu.mo)
}
}

% The paper headers
\markboth{Journal of \LaTeX\ Class Files,~Vol.~XX, No.~X, XX~XXXX}%
{Shell \MakeLowercase{\textit{et al.}}: A Sample Article Using IEEEtran.cls for IEEE Journals}

% \IEEEpubid{0000--0000/00\$00.00~\copyright~XXXX IEEE}
% Remember, if you use this you must call \IEEEpubidadjcol in the second
% column for its text to clear the IEEEpubid mark.

\maketitle

\begin{abstract}
Gaussian Splatting demonstrates impressive results in multi-view reconstruction based on Gaussian explicit representations. However, the current Gaussian primitives only have a single view-dependent color and an opacity to represent the appearance and geometry of the scene, resulting in a non-compact representation.
In this paper, we introduce a new method called \textbf{SVGS} (Spatially Varying Gaussian Splatting) that utilizes spatially varying colors and opacity in a single Gaussian primitive to improve its representation ability. 
We have implemented bilinear interpolation, movable kernels, and tiny neural networks as spatially varying functions. 
\XR{\textbf{SVGS} employs 2D Gaussian surfels as primitives, which significantly enhances novel-view synthesis while maintaining high-quality geometric reconstruction. This approach is particularly effective in practical applications, as scenes combining complex textures with relatively simple geometry occur frequently in real-world environments.}
Quantitative and qualitative experimental results demonstrate that all three functions outperform the baseline, with the best movable kernels achieving superior novel view synthesis performance on multiple datasets, highlighting the strong potential of spatially varying functions. 
\end{abstract}


\section{Introduction}
\label{sec:intro}

% \XR{ToDO:
% add PGSR CD value limit GS number.
% time compare with 2dgs.
% vs Textured-GS.
% mem use in training and rendering.
% diff MLP.
% move tables.
% vs mip-splatting, zip nerf.
% Unreasonable negative values
% }

Novel-view synthesis (NVS) has always been an important task in computer graphics and computer vision, with various applications in robotics, AR/VR, and autonomous driving.
Compared with neural radiance fields (NeRF)~\cite{mildenhall2021nerf}-based methods, recent Gaussian splatting methods~\cite{3dgs,2dgs, xu2024texture, huang2024textured, chao2024textured,fan2025lightgaussian, fang2024mini, lee2024compact, mallick2024taming} directly reconstruct 3D scenes by splatting explicit Gaussian primitives like ellipsoids~\cite{3dgs} or surfels~\cite{2dgs}, achieving significant progress in novel view synthesis and geometric reconstruction.

Though impressive NVS quality has been achieved by these Gaussian splatting-based methods, these methods are ineffective and non-compact in representing a complex scene.
% \ly{todo: how it works, why not effective, an example (refer to teaser)}.
In these methods, the input images are fitted by splatting a set of Gaussian primitives. Each primitive only has a single view-dependent color and an opacity to represent the appearance and geometry of the scene. However, when the scene has complex geometry and appearance, these methods have to create a large number of these simple Gaussians to approximate the spatially varying opacity and textures on the scene, which leads to a huge waste of Gaussians. 
% \XR{Existing Gaussian splatting-based methods~\cite{3dgs,2dgs} use a large number of Gaussian primitives to represent the scene. 
% However, each Gaussian primitive only represents a single color and opacity, making it challenging to reconstruct high-frequency detail images as shown in Fig.~\ref{fig:teaser}.}
%
\begin{figure}[!tp]
    \centering
    \begin{overpic}[width=\linewidth]{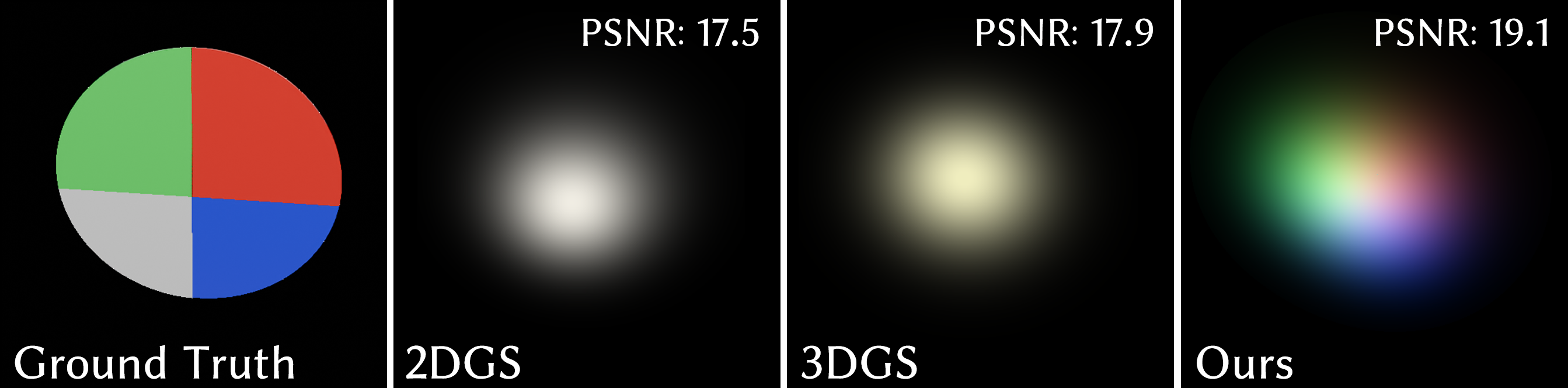}
    % \put(9,  -1.7){2DGS}
    % \put(34, -1.7){3DGS}
    % \put(58, -1.7){Ours-MK}
    % \put(82, -1.7){Ground Truth}
    \end{overpic}
    \vspace{-8mm}
    \caption{
    \XR{A toy example on single Gaussian fitting. The ground-truth input image is a four-color disk, and the training is restricted to using only one single Gaussian. Under this constraint, both 2DGS~\cite{2dgs} and 3DGS~\cite{3dgs} produce monochromatic outputs, whereas \name~can represent significantly richer color variations using just one Gaussian primitive.}
    }
\label{fig:single}
\vspace{-4mm}
\end{figure}

\begin{figure*}[!tp]
    \centering
    \begin{overpic}[width=\linewidth]{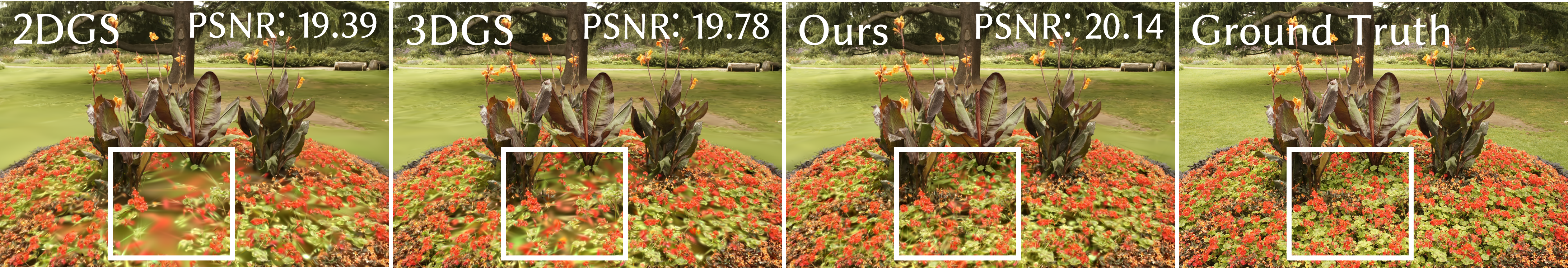}
    % \put(9,  -1.7){2DGS}
    % \put(34, -1.7){3DGS}
    % \put(58, -1.7){Ours-MK}
    % \put(82, -1.7){Ground Truth}
    \end{overpic}
    \vspace{-8mm}
    \caption{
    \XR{\name~gives each Gaussian the ability to vary spatially. Compared with 2DGS~\cite{2dgs} and 3DGS~\cite{3dgs}, \name~is more expressive and can better reconstruct details (as seen in the white area). }
    }
\label{fig:teaser}
% \vspace{-4mm}
\end{figure*}

To address this problem, we introduce a new method called \textbf{SVGS} (Spatially Varying Gaussian Splatting) that utilizes spatially varying colors and opacity in a single Gaussian primitive to improve its representation ability. 
% \ly{todo: briefly introduce the meaning of spatially varying, why it is better than vanilla one, examples in teaser.}
% \XR{
% We define the intersection point between each ray and the current Gaussian primitive. 
% As the position of this intersection point changes, the color and opacity values returned by this Gaussian will also change. 
% This makes a single Gaussian more capable of fitting complex color scenes, increases the expressive capability, reduces waste, and makes the representation of complex scenes more compact and effective, as shown in Fig.~\ref{fig:teaser}.
% }
This spatially varying attribute means that different rays that intersect the same Gaussian primitive may have different colors if these rays intersect the Gaussian at different locations. 
\XR{An example is shown in Fig.~\ref{fig:single}, where our target is to fit a round plane with four different colors using only one single Gaussian primitive, while the original 2DGS~\cite{2dgs} or 3DGS~\cite{3dgs} all fail to reconstruct the colors of this simple shape by using one Gaussian primitive.}
In the vanilla Gaussian Splatting, Gaussian primitives always have the same opacity or view-dependent colors for all rays while our spatially varying Gaussians show different colors for different intersection points.
This makes a single Gaussian more capable of fitting complex textures and geometry in the scene, increasing representation ability and making our representation more compact and effective, as shown in Fig.~\ref{fig:teaser}.

% For each Gaussian surfel, different locations have different colors and opacity, as shown in the toy example in the bottom right of Fig.~\ref{fig:teaser} and Fig.~\ref{fig:func}. 
% Thus, each Gaussian surfel can express more color information than the vanilla one, improving our overall scene representation and reconstruction capabilities.}

% \ly{todo: introduce all three types of spatially varying functions and discuss their pros and cons \textit{theoretically}.}
% \XR{
\XR{The fundamental distinction between our method and the vanilla Gaussian Splatting framework is shown in Fig.~\ref{fig:func}.
To define the spatially varying function inside a Gaussian primitive, we try three different designs. 
All three spatially varying functions are implemented based on Gaussian surfels~\cite{2dgs}.}
The first function divides each Gaussian surfel into four quadrants using bilinear interpolation, assigning a learnable color and opacity value to each quadrant, which enhances color expression but may cause gradient vanishing issues (See Fig.~\ref{fig:2func} (a)).
In the second design, we define four movable kernels based on the original Gaussian surfel,  providing higher flexibility and stronger expressiveness, as shown in Fig.~\ref{fig:2func} (b).
Third, we apply a tiny three-layer neural network on each Gaussian surfel that can return a color and opacity value for any intersection point on the surfel. Such neural network-based representation shows strong representation ability but with significantly more parameters than the other two functions.

% }
\XR{SVGS adopts 2D Gaussian surfels as its primitive representation, substantially improving novel-view synthesis performance while maintaining geometric reconstruction quality, particularly in cases where textures are complex but the underlying geometry is simple (as in the Blender~\cite{mildenhall2021nerf} dataset). Such scenarios are very common in real-world environments, and our approach also generalizes well to scenes with more intricate geometric structures.}
To demonstrate the effectiveness of \name, we conduct experiments on the Synthetic Blender~\cite{mildenhall2021nerf}, DTU~\cite{jensen2014DTU}, Mip-NeRF360~\cite{barron2022mipnerf360}, and Tanks\&Temples~\cite{knapitsch2017tanks} datasets to validate all three designs. 
Experimental results demonstrate that all three spatially varying functions outperform the baseline method 2DGS~\cite{2dgs} in novel view synthesis, while the compact movable kernels design achieves the best results. 
\XR{Moreover, \name~with movable kernels surpasses all other Gaussian Splatting-based methods on the Blender~\cite{mildenhall2021nerf} dataset, demonstrating its ability to represent complex textures on relatively flat geometries.
}
We further demonstrate the compactness of \name~by using a limited number of Gaussian primitives \XR{and limited training times} to achieve superior rendering quality.

\section{Related Works}
\label{sec:relatedworks}

3D reconstruction has been widely studied in the past. 
Different from reconstructing geometric models from point clouds~\cite{kazhdan2006poisson,nan2017polyfit,xu2022rfeps,lin2022surface,xu2023globally}, reconstructing images~\cite{seitz2006comparison,schonberger2016structure,mildenhall2021nerf,3dgs} and shapes~\cite{wang2021neus,2dgs,guedon2024sugar} from multi-view images has always been a harder problem to be solved.

\subsection{Novel View Synthesis}
Seitz et al.~\cite{seitz2006comparison} introduces multi-view stereo (MVS) reconstruction algorithms that determine per-view depth maps by maximizing multi-view consistency through patch or feature-level matching, followed by surface reconstruction via multi-view fusion. 
Subsequent methods like COLMAP~\cite{schoenberger2016colmap}, OpenMVS~\cite{openMVS}, and PMVS~\cite{furukawa2010accurate} excel on texture-rich, flat surfaces but struggle in textureless areas and near occlusion boundaries. 
Recently, learning-based MVS approaches, such as MVSNet~\cite{yao2018mvsnet} and its variants~\cite{yao2019recurrent, luo2019pmvsnet, yu2020fast, zhang2023vis}, have mitigated the issues in textureless regions, 
but suffer from lack of multi-view consistency due to the independent depth prediction for each view.
Schonberger et al.~\cite{schonberger2016structure} proposes a incremental Structure-from-Motion (SfM) technique that addresses key challenges in robustness, accuracy, completeness, and scalability.
The famous NeRF~\cite{mildenhall2021nerf} and its variants~\cite{barron2021mip, barron2022mip, barron2023zip, huang2023nerf} presents a method that achieves remarkable results in synthesizing novel views of complex scenes by optimizing a continuous volumetric scene function with a sparse set of input views using a fully connected deep network, effectively rendering photorealistic images through differentiable volume rendering.

\subsection{Gaussian-based methods.}
Recently, 3DGS~\cite{3dgs} achieves impressive visual quality and real-time novel-view synthesis by using 3D Gaussians for scene representation, interleaved optimization of anisotropic covariance, and a fast visibility-aware rendering algorithm, demonstrating superior results on several established datasets. 

Due to the excellent explicit expression ability of 3DGS~\cite{3dgs}, many methods based on Gaussian splatting have been proposed.
Scaffold-GS~\cite{lu2024scaffold} uses anchor points to distribute local 3D Gaussians and dynamically predicts their attributes based on viewing direction and distance, reducing redundant Gaussians and improving scene coverage, thereby enhancing rendering quality.
Mip-splatting~\cite{yu2024mip} introduces a 3D smoothing filter to constrain the size of 3D Gaussian primitives and a 2D Mip filter to mitigate aliasing and dilation issues, demonstrating effectiveness across multiple scales. 
GES~\cite{hamdi2024ges} improves 3D scene representation efficiency and accuracy over Gaussian Splatting by using fewer particles and a frequency-modulated loss, significantly reducing memory footprint and increasing rendering speed. 
3D-HGS~\cite{li20243d} addresses the limitations of 3DGS~\cite{3dgs} in representing discontinuous functions, demonstrating improved performance and rendering quality without compromising speed. 
\FF{Splat-the-Net~\cite{zhou2025splat} further enhances primitive expressivity by representing each splattable primitive as a bounded neural density field parameterized by a shallow neural network. By deriving an exact analytical solution for line integrals, it enables perspectively accurate splatting without costly ray marching, achieving comparable rendering quality and speed to 3DGS while using significantly fewer primitives and parameters.}
PixelSplat~\cite{charatan2024pixelsplat} introduces a feed-forward model that reconstructs 3D radiance fields from image pairs using 3D Gaussian primitives, achieving significantly faster 3D reconstruction on novel view synthesis.
\XR{Texture-GS~\cite{xu2024texture} disentangles appearance from geometry in 3D Gaussian Splatting by learning UV mappings and applying 2D textures to 3D Gaussians, enabling flexible appearance editing such as texture swapping while maintaining high-fidelity reconstruction and real-time rendering performance.
Textured-GS~\cite{huang2024textured} extends Gaussian Splatting by introducing spatially defined color and opacity through spherical harmonics, allowing each Gaussian to represent richer appearance variations without increasing primitive count and thereby improving rendering fidelity.
Textured-Gaus~\cite{chao2024textured} enhances 3D Gaussian Splatting by equipping each Gaussian with alpha and RGB texture maps to model spatially varying color and opacity, significantly boosting the expressivity and rendering quality of individual primitives while maintaining efficient reconstruction and rendering performance.
MCMC-3DGS~\cite{kheradmand20243d} reinterprets 3D Gaussian Splatting as a Markov Chain Monte Carlo sampling process, replacing heuristic cloning and splitting with principled stochastic updates via Stochastic Gradient Langevin Dynamics, thereby improving rendering quality, initialization robustness, and controllability over the number of Gaussians.}
Following 3DGS~\cite{3dgs}, 2DGS~\cite{2dgs}, Gaussian Surfels~\cite{Dai2024GaussianSurfels} and Gaussian billboards~\cite{weiss2024gaussian} was proposed by compressing the ellipsoid into a Gaussian surfel by defining the shortest axis of the ellipsoid as the normal vector, achieving high-quality geometric reconstruction while retaining the ability to reconstruct from novel views.
\XR{PGSR~\cite{chen2024pgsr} further advances Gaussian-based reconstruction by introducing a planar-based Gaussian splatting framework that explicitly models local surface geometry. It employs an unbiased depth rendering strategy and multi-view geometric regularization to enhance global consistency, achieving high-fidelity surfaces and photorealistic rendering with fast optimization and inference.
}

\section{Method}
\label{sec:method}

\subsection{Spatially Varying Gaussian Primitives}
\textbf{Gaussian Splattings.}
Given multi-view images with the corresponding camera poses, our target is to render novel-view images.
We achieve this by representing the whole scene with a set of trainable Gaussian primitives. Then, to train these parameters of all Gaussian primitives, we apply the splatting technique to render images on the input viewpoints and minimize the difference between the rendered images and input images. After learning the Gaussian primitives, we can apply the same splatting technique to render images of arbitrary viewpoints.

% 3D Gaussian Splatting~\cite{3dgs} uses a set of Gaussian ellipsoids in the world space to represent the whole scene. 
% The Gaussian function for the point $p$ is defined as a Gaussian function $G(p)$.
 % = e^{-\frac{1}{2}(p_i)^T\sum^{-1}(p_i)}
% where the matrix $\sum$ is defined as the three-dimensional covariance matrix in world space.
% A color value $\mathbf{c}_{SH}(d)\in\mathbf{R}^3$ and a opacity value $\alpha$ are defined on each Gaussian ellipsoid, and the color here is defined by the spherical harmonic function which related to the ray incident direction $d$.
% After we get the actual color $\mathbf{c}_{SH}(d)$, we can perform $\alpha$-blending based on its opacity $\alpha$ and the Gaussian function value to get a final color on each pixel.
% and they use the values of the Gaussian function as coefficients for the subsequent $\alpha$-blending.

\textbf{Colors and Opacity of Gaussian Primitives.} 
The colors and opacity of Gaussian primitives in existing methods~\cite{3dgs,2dgs,li20243d,yu2024mip} are independent of the intersection locations with the primitives, which leads to ineffective representation ability for complex textures or geometry in a complex scene.
% In all these methods, each Gaussian primitive contains a set of attributes, including position, scale, orientation, colors, and opacity.
% Among all these attributes, the colors define the appearance while the opacity affects the underlying geometry.
The colors are usually represented by a view-dependent spherical harmonic function $SH(\mathbf{d})$~\cite{muller2022instant, fridovich2022plenoxels}, where $\mathbf{d}$ is the viewing ray direction from the current pixel.
The opacity $\alpha$ is a single value associated with the Gaussian primitive.
To render an image, Gaussian splatting methods perform alpha-blending~\cite{zwicker2001ewa} under these Gaussian primitives. 
A noticeable issue is that the Gaussian primitives report exactly the same color with the same viewing direction but intersect this primitive with different intersection points. 
Utilizing such color function forces to represent an underlying surface with complex textures leads to ineffective and redundant small Gaussian primitives.
The same issue also exists for the opacity, which has difficulty in representing complex textures and geometries.

% A color value  and an opacity value 
% $\alpha$ are defined on each Gaussian ellipsoid, where the color is determined by the spherical harmonic function related to the ray incident direction $d$. After obtaining the actual color 
% $\mathbf{c}_{SH}(d)$, we perform 
% $\alpha$-blending~\cite{zwicker2001ewa} based on its opacity 
% $\alpha$ and the Gaussian function value to get a final color for each pixel.
% The recently introduced 2D Gaussian Splatting~\cite{2dgs} flattens the Gaussian ellipsoid into a planar surfel, like an ellipse, and defines the axis of the steepest change of density as the normal vector at this point. They use similar coordinate transformations and 
% $\alpha$-blending mechanisms to achieve fast reverse rendering.
% However, the color of each primitive in 2DGS~\cite{2dgs} and 3DGS~\cite{3dgs} is only related to the direction 
% $d$ of the incident ray. Regardless of where the light intersects the Gaussian primitive, their spherical harmonics will return the same color value. To address this, we propose a spatial variation function based on the intersection position 
% $p$ with the primitive for further improvement.

\begin{figure}[!tp]
    \centering
    \begin{overpic}[width=0.98\linewidth]{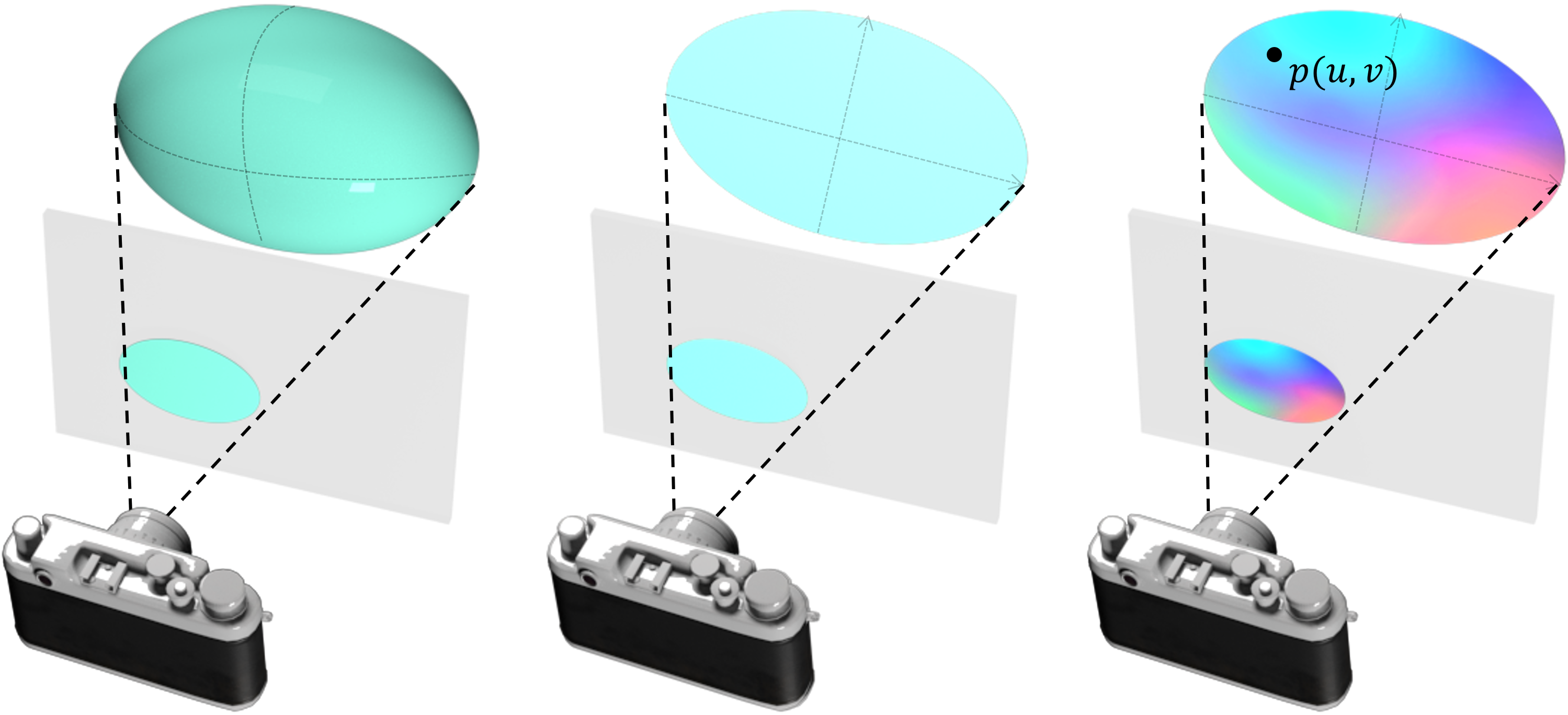}
    % \put(-1.5, 32){(a) Ours w/o Decay}
    % \put(96, 17){$\mathbf{u}$}
    % \put(77, 30){$\mathbf{v}$}
    \put(6,  -4.0){(a) 3DGS}
    \put(38, -4.0){(b) 2DGS}
    \put(69, -4.0){(c) \name}
    \end{overpic}
    \vspace{2pt}
    \caption{
    3DGS~\cite{3dgs} uses Gaussian ellipsoids to express scenes, and a learnable color is defined on each ellipsoid. 2DGS~\cite{2dgs} uses Gaussian surfels to express scenes, and a learnable color is defined on each Gaussian surfel. Our \name~uses spatially varying Gaussian surfels to express scenes, and the color and opacity changes with the spatial position on each surfel.
    }
\label{fig:func}
% \vspace{-4mm}
\end{figure}

\textbf{Spatially Varying Colors and Opacity.}
To address the above issue, we propose spatially varying colors and opacity in \name. Specifically, we use a color function $\mathbf{c}(\mathbf{p},\mathbf{d})$ and an opacity function $\alpha(\mathbf{p})$ as
\begin{equation}
    \mathbf{c}(\mathbf{p},\mathbf{d})=SH(\mathbf{d})+\mathcal{F}_\mathbf{c}(\mathbf{p}),
\end{equation}
\begin{equation}
    \alpha(\mathbf{p})= \mathcal{F}_\alpha(\mathbf{p}),
\end{equation}
where $\mathbf{p}$ is the intersection point between the given Gaussian primitive and the ray from current pixel (Fig.~\ref{fig:func} (c)). 
\XR{$\mathcal{F}_\mathbf{c}(\mathbf{p})$ and $\mathcal{F}_\alpha(\mathbf{p})$ denote the spatially varying functions for color and opacity, respectively. By defining the opacity and colors through spatially varying functions, we significantly enhance the representational power of Gaussian primitives, allowing them to better capture complex textures and geometric variations. We do not impose explicit constraints on the values of $\mathcal{F}_{\mathbf{c}}$ and $\mathcal{F}_{\alpha}$, so these functions may take negative values. The negative outputs are permissible because they contribute additively to the original spherical harmonic color representation. This design choice is justified by the fact that, during both training and rendering in Gaussian splatting~\cite{3dgs}, the final color and opacity are normalized to the valid range via a sigmoid activation. Enforcing additional value constraints during optimization would be non-trivial and computationally expensive; hence, we allow unconstrained intermediate values while relying on the activation to ensure physically meaningful results.
Moreover, the optimization framework automatically removes Gaussians whose opacity falls below a critical threshold.
}
Thus, recent works like 3D-HGS~\cite{li20243d} and NegGS~\cite{kasymov2024neggs} are special cases of our SVGS. 
% By defining the opacity and colors with spatially varying functions, we greatly improve the representation ability and make the Gaussian primitives more effective for complex textures and geometry. And we do not pose any constraints on the values of $F_\mathbf{c}$ and $F_{\alpha}$ so these functions could be negative. 
% The negative values are permissible as they contribute additively to the original spherical harmonic colors. 
% \XR{This is because during the training and rendering of Gaussian splatting, the final color is scaled to the available value range through a sigmoid activation function, and adding value constraints to individual variables within the optimization process is difficult and computationally expensive. Therefore, we do not explicitly constrain them.}

% \ly{maybe, GES and other functions are also special cases of our method.}
% \XR{I think is a special case of Spatially varying function, but not our method in this paper?}

\textbf{Computation of Intersection Point $\mathbf{p}$.} To compute an intersection point for our method, we adopt the 2D Gaussian Splatting~\cite{2dgs} to use surfels as the Gaussian primitives. Then, the intersection point $\mathbf{p}$ is defined as the intersection point on the Gaussian surfel $\mathbf{p}=(u,v)$ (Fig.~\ref{fig:func} (c)). 
The coordinates of the intersection point $\mathbf{p}$ are defined in the local 2D coordinate system of the Gaussian ellipse, where the Gaussian origin point serves as $(0, 0)$ and the ellipse axes form the coordinate axes.
Note that we use 2DGS by default for simplicity, but our discussion can also be extended to 3D Gaussians by regarding the 3D Gaussians as ellipsoids and using the intersection points with the ellipsoids, while the calculation of intersection points requires careful consideration.
In the following, we discuss three different implementations of our spatially varying functions $\mathcal{F}_\mathbf{c}$ and $\mathcal{F}_{\alpha}$.

\begin{figure}[!tp]
    \centering
    \begin{overpic}[width=0.98\linewidth]{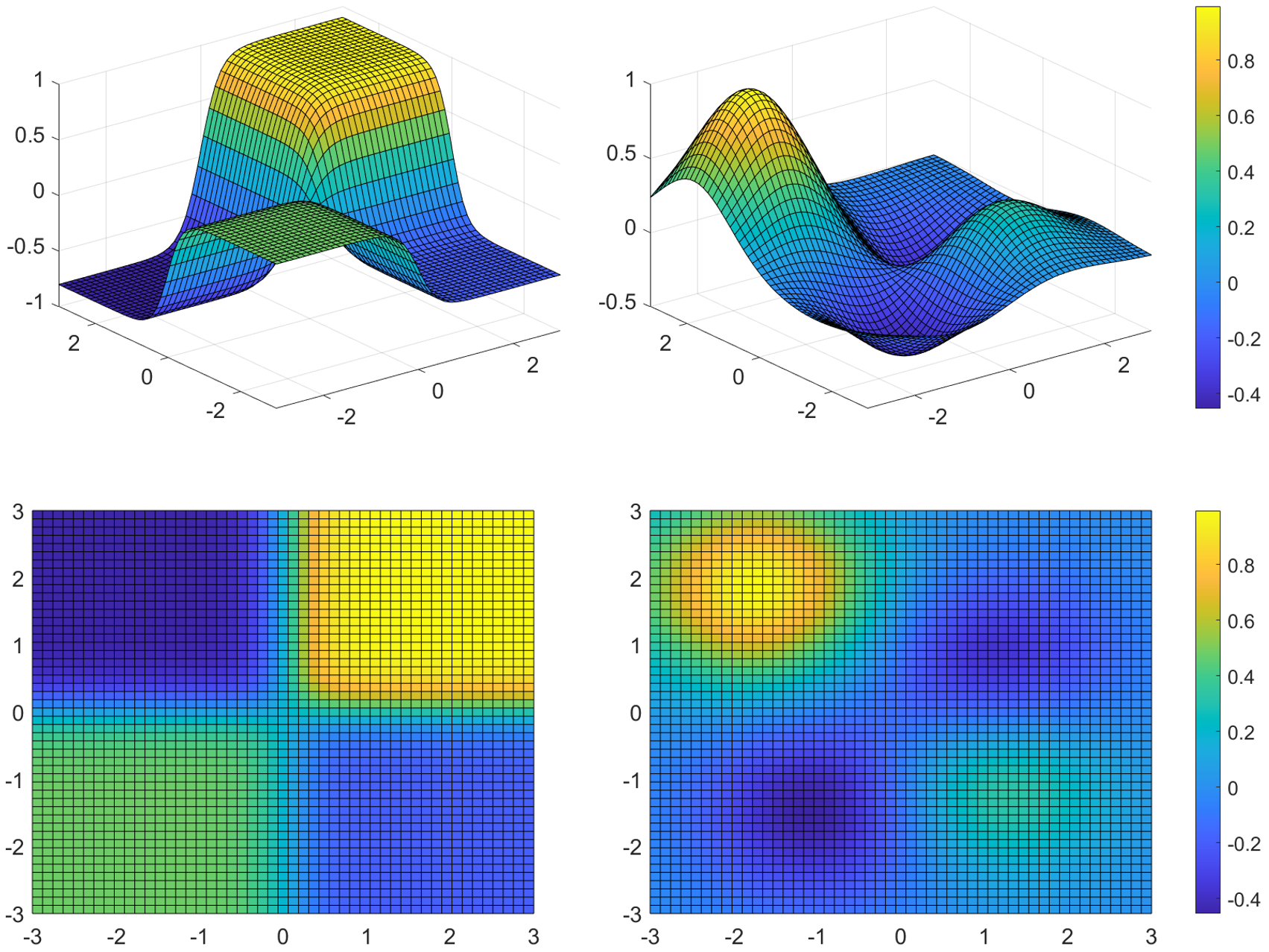}
    % \put(-1.5, 32){(a) Ours w/o Decay}
    % \put(96, 17){$\mathbf{u}$}
    % \put(77, 30){$\mathbf{v}$}
    \put(0, 37){(a) Bilinear Interpolation}
    \put(54, 37){(b) Movable kernels}
    \put(0, -3.2){(c) Bilinear Interpolation}
    \put(54, -3.2){(d) Movable kernels}
    \end{overpic}
    \vspace{0pt}
    \caption{
    Visualization of the bilinear interpolation (a,c) and movable kernel functions (b,d).
    }
\label{fig:2func}
% \vspace{-4mm}
\end{figure}

\subsection{Bilinear Interpolation}
% First of all, for a function on a two-dimensional plane, bilinear interpolation is a common choice to split it up. 
In this function, we use bilinear interpolation to divide each elliptical Gaussian into four quadrants, where each quadrant has different color and opacity values. 
Then calculate $\mathbf{c}(\mathbf{p}, \mathbf{d})$ and $\alpha(\mathbf{p})$ at any position in object space through bilinear interpolation.

The bilinear interpolated color can be obtained by a simple bilinear interpolation
\begin{equation}
\begin{aligned}
    \mathcal{F}_\mathbf{c}(\mathbf{p}) &= (1- u')  (1 - v') \mathbf{c}_{0} +  (1- u')  v' \mathbf{c}_{1}
    \\&+  u'  (1 - v') \mathbf{c}_{2} +  u' v' \mathbf{c}_{3},
\end{aligned}
\end{equation}
and the same goes for opacity
\begin{equation}
\begin{aligned}
    \mathcal{F}_\mathbf{\alpha}(\mathbf{p}) &= (1- u')  (1 - v') \alpha_{0} +  (1- u')  v' \alpha_{1}
    \\&+  u'  (1 - v') \alpha_{2} +  u' v' \alpha_{3},
\end{aligned}
\end{equation}
where $\mathbf{c}_i$ and $\alpha_i$ for \(i = 0, 1, 2, 3\) are the four new learnable colors and opacities corresponding to the four quadrants.
We also use a simple sigmoid function to rescale the coordinates \(\mathbf{p} = (u, v)\) in the object space to \((0, 1)\) to avoid some irregular values:
\begin{equation}
    u' = \frac{1}{1 + e^{-\lambda_s u}}, \quad v' = \frac{1}{1 + e^{-\lambda_s v}},
\label{eq:sigmoid}
\end{equation}
where \(\lambda_s\) is the parameter that controls changing rate of sigmoid function. We set it to 5.0 by default.
% The degree of transition between different quadrants is controlled by $\lambda_s$, 
% we show the influence of $\lambda_s$ in Appendix~\ref{Asec:lambads}.

% More specifically, for each Gaussian surfel $p_i$, we define four new learnable colors $\mathbf{c}_{i,j} = (c_{i,j}^r,c_{i,j}^g,c_{i,j}^b)$ and opacities $\alpha_{i,j}$ on each Gaussian, where $j={0,1,2,3}$ 

\begin{figure}[!tp]
    \centering
    \begin{overpic}[width=0.98\linewidth]{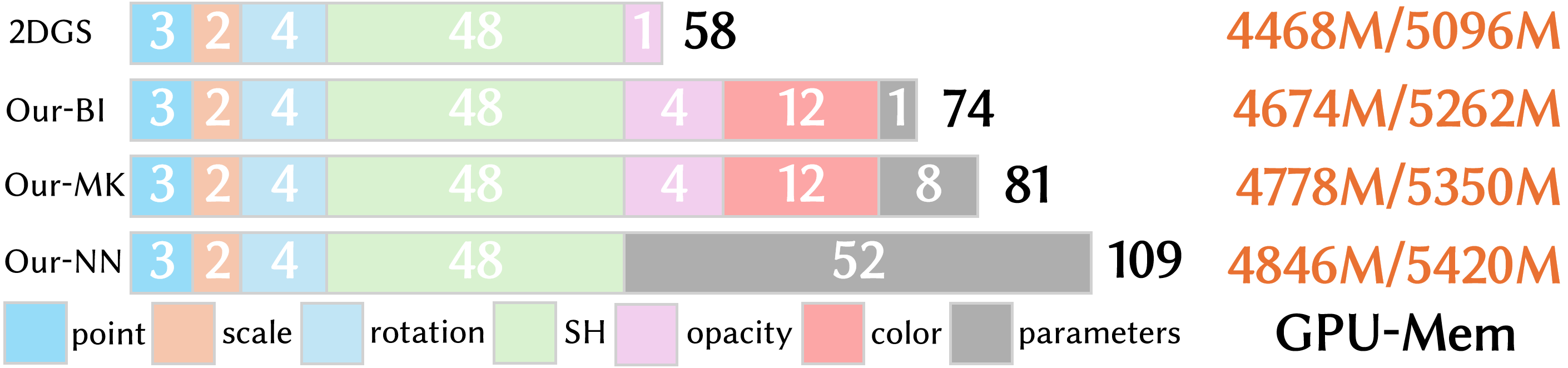}
    % \put(-1.5, 32){(a) Ours w/o Decay}
    % \put(96, 17){$\mathbf{u}$}
    % \put(77, 30){$\mathbf{v}$}
    % \put(0, 37){(a) Bilinear Interpolation}
    % \put(54, 37){(b) Movable kernels}
    % \put(0, -3.2){(c) Bilinear Interpolation}
    % \put(54, -3.2){(d) Movable kernels}
    \end{overpic}
    \vspace{-3mm}
    \caption{
    The parameter amounts of the three proposed spatial variation functions and the original 2DGS~\cite{2dgs} on a single Gaussian. The parameter amounts of the three proposed functions are 1.28 times, 1.40 times, and 1.88 times of the original 2DGS~\cite{2dgs}. \XR{And average GPU memory useage (in MB) during training and rendering.}
    }
\label{fig:paras}
% \vspace{-4mm}
\end{figure}

\subsection{Movable Kernels}
The above bilinear interpolation method can be regarded as four fixed kernels located in the four quadrants of the elliptical Gaussian surfel. 
This inspires us to further enhance its expressiveness by using movable kernels.

We define \(k\) movable kernels \(\mathbf{K}_i = (K^x_i, K^y_i)\) on each Gaussian surfel, where \(i = 0, 1, 2, \ldots, k-1\) corresponds to the index of kernels. Assume that the intersection point of the current pixel and Gaussian surfel is \(\mathbf{p} = (u, v)\). The color and opacity can be calculated by:
\begin{equation}
\begin{aligned}
\mathcal{F}_\mathbf{c} &= \sum_{i=0}^{k-1} \mathcal{F}_{\mathbf{K}_i}(\mathbf{p}) \mathbf{c}_i, \\
\mathcal{F}_\mathbf{\alpha} &= \sum_{i=0}^{k-1} \mathcal{F}_{\mathbf{K}_i}(\mathbf{p}) \alpha_i,
\end{aligned}
\end{equation}

In our implementation, each kernel is represented as a separate exponential function that decays as the distance from the point \(\mathbf{p}\) to the kernel center \(\mathbf{K}_i\)
\begin{equation}
\mathcal{F}_{\mathbf{K}_i}(\mathbf{p}) = e^{-\lambda_e \left\| \mathbf{p} - \mathbf{K}_i \right\|^2},
\end{equation}
where \(\lambda_e\) is similar to the \(\lambda_s\) mentioned in the previous section, both of which are used to control the changing rate of the kernel function. We set \(\lambda_e = 0.1\) and \(k = 4\) by default. 
Fig.~\ref{fig:2func} demonstrates the numerical visualizations of these two spatially varying functions, and we can also choose other kernel functions like the sigmoid function as introduced in Sec.~\ref{sec:results}.

% We also show ablation experiments on different kernel functions in Sec.~\ref{sec:results}, by replacing the exponential function with the sigmoid function or using more kernels.

% \begin{equation}
% \begin{aligned}
% \mathbf{c}_{MK} &= \sum_{j=0}^k e^{-\lambda_k[(u_i - k^x_{i,j})^2 +(v_i - k^y_{i,j})^2] } \mathbf{c}_{i,j} \\
% \alpha_{MK} &= \sum_{j=0}^k e^{-\lambda_k[(u_i - k^x_{i,j})^2 +(v_i - k^y_{i,j})^2] } \alpha_{i,j}
% \end{aligned}
% \end{equation}
% Similarly, we directly use the opacity $\alpha_{MK}^i$ we calculated, and then add the new color $\mathbf{c}_{MK}^i$ to the color obtained by the original spherical harmonic function $\mathbf{c}_{SH}^i$.
% \begin{equation}
%     \mathbf{c}^i = \mathbf{c}_{SH}^i + \mathbf{c}_{MK}^i.
% \end{equation}

\subsection{Tiny MLPs}
\begin{wrapfigure}{r}{3cm}
\vspace{-3.5mm}
  \hspace*{-1.5mm}
  \centerline{
  \includegraphics[width=35mm]{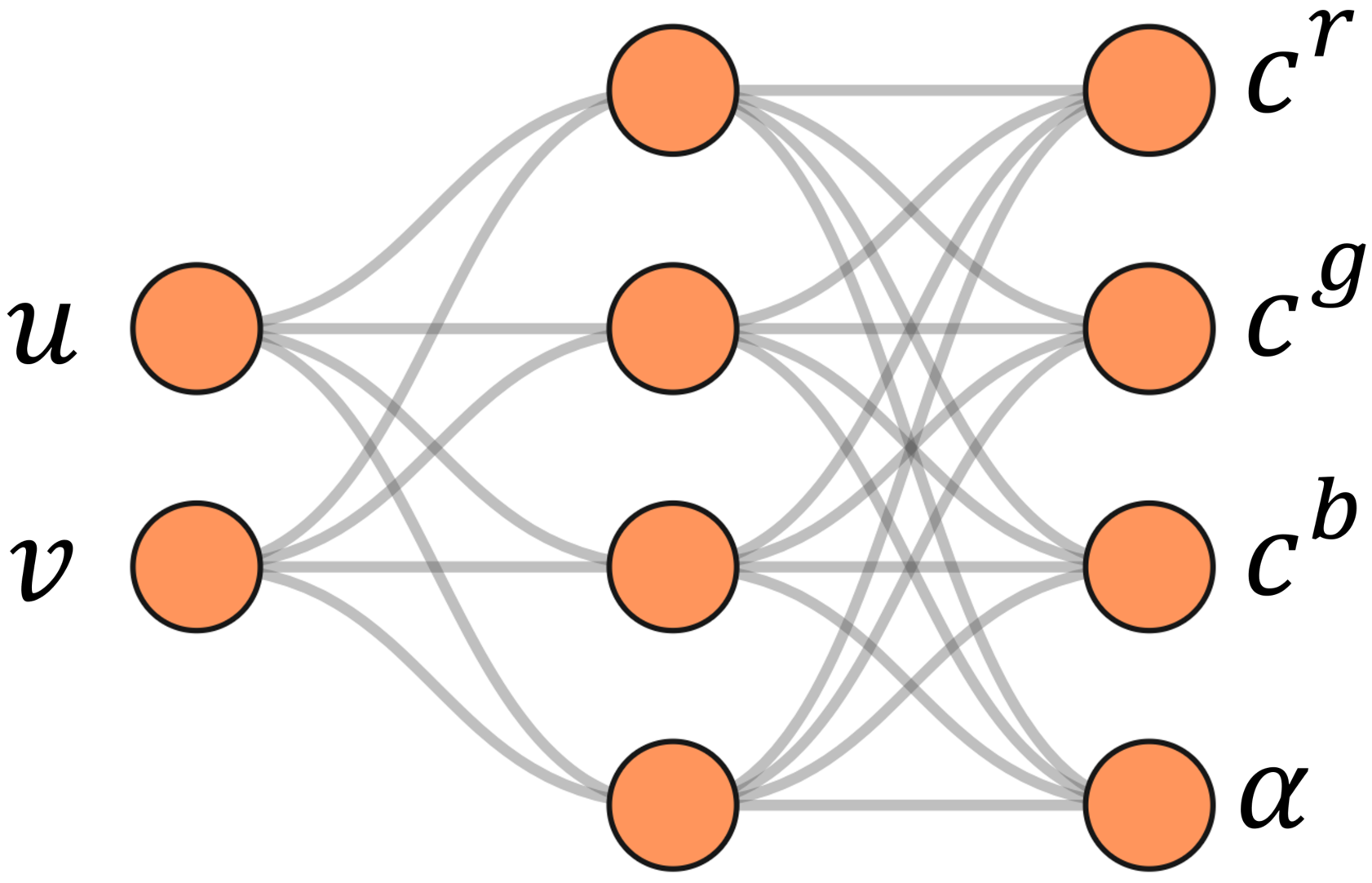}
  }
\centering
\vspace*{-6mm}
  \caption{
    \XR{MLP architecture.}
    }
\label{fig:mlp}
  \vspace*{-6mm}
\end{wrapfigure}
% In order to prove that the designed kernel function has the characteristics of enhanced expressiveness and ease of training, we also tried to use a tiny neural network to fit the colors on each primitive. 
Instead of using an interpolation or kernel function to represent the spatially varying function, we define a separate small multilayer perceptron (MLP) on each Gaussian surfel. To reduce the number of parameters as much as possible, we only use a tiny MLP with a three-layer network taking local coordinates $(u, v)$ as input and producing RGB colors with opacity values as the spatial variation function (as shown Fig.~\ref{fig:mlp})
\begin{equation}
    \mathcal{F}_\mathbf{c}, \mathcal{F}_\mathbf{\alpha} = \text{MLP}(\mathbf{p}),
\end{equation}
where the MLP accepts a two-dimensional input \(\mathbf{p} = (u, v)\) and outputs the color and opacity at that location, as shown in the wrapped figure. Inside the MLP, we adopt the sigmoid function is used as the activation function. 
% Each MLP layer is defined by a matrix and a bias, and the sigmoid function is used as the activation function. 
However, although we adopt a shallow three-layer MLP, the number of parameters still far exceeds the other two functions, as shown in Fig.~\ref{fig:paras}.
\XR{A detailed ablation in Sec.~\ref{sec:ablation} analyzes how the number of MLP layers affects the final reconstruction quality.}

% Fig.~\ref{fig:2func} shows visualizations of two spatially varying functions: bilinear interpolation and movable kernels, while the tiny MLP cannot be explicitly expressed.
% See Appendix~\ref{Asec:DissOnSVF} for more detailed discussion on these three different spatially varying functions.

% Fig.~\ref{fig:diffff} demonstrates the image fitting capabilities of these three spatial variation functions on a single Gaussian. Although bilinear interpolation can better fit color mutations, it suffers from the problem of gradient vanishing in other areas.

\section{Experimental Results}
\label{sec:results}

\begin{table*}[!tp]
    \centering
    \caption{Comparisons of three different spatially varying functions and the original 2DGS~\cite{2dgs} with different number limits. 
This table shows the PSNR, SSIM and LPIPS metrics on the Synthetic Blender dataset~\cite{mildenhall2021nerf}, Mip-NeRF360~\cite{barron2022mipnerf360} dataset, Tanks\&Temples~\cite{knapitsch2017tanks} dataset, and DTU~\cite{jensen2014DTU} dataset. 
The top three results are highlighted in \sethlcolor{rk1}\hl{red}, 
\sethlcolor{rk2}\hl{orange}, and \sethlcolor{rk3}\hl{yellow}, respectively. The notation w/o. \(\mathcal{L}_N\) indicates that no normal loss is applied. 
}
\vspace{-2mm}
    \resizebox{\linewidth}{!}{
\begin{tabular}{l|cccccc|cccccc|cccccc|cccccc}
\toprule
Dataset & \multicolumn{6}{c|}{Synthetic Blender}                                                                                           & \multicolumn{6}{c|}{Mip-NeRF360}                                                                                                  & \multicolumn{6}{c|}{Tanks\&Temples}                                                                                              & \multicolumn{6}{c}{DTU}                                                           \\ \midrule
GS Num  & \multicolumn{3}{c|}{50K}                                                  & \multicolumn{3}{c|}{noLimit}                         & \multicolumn{3}{c|}{500K}                                                  & \multicolumn{3}{c|}{noLimit}                         & \multicolumn{3}{c|}{500K}                                                 & \multicolumn{3}{c|}{noLimit}                         & \multicolumn{3}{c|}{100K}                & \multicolumn{3}{c}{noLimit}             \\ \midrule
Metrics  & \multicolumn{1}{l}{PSNR}              & SSIM & \multicolumn{1}{l|}{LPIPS} & \multicolumn{1}{l}{PSNR}              & SSIM & LPIPS & \multicolumn{1}{l}{PSNR}               & SSIM & \multicolumn{1}{l|}{LPIPS} & \multicolumn{1}{l}{PSNR}              & SSIM & LPIPS & \multicolumn{1}{l}{PSNR}              & SSIM & \multicolumn{1}{l|}{LPIPS} & \multicolumn{1}{l}{PSNR}              & SSIM & LPIPS & \multicolumn{1}{l}{PSNR} & SSIM & \multicolumn{1}{l|}{LPIPS} & \multicolumn{1}{l}{PSNR} & SSIM & LPIPS \\ \midrule
2DGS    & 32.64                                 &      \cellcolor{rk3}0.963& \multicolumn{1}{l|}{\cellcolor{rk3}0.042}      & 32.87                                 &      0.965&       0.038& 25.93                                  &      0.755& \multicolumn{1}{l|}{0.314}      & 26.86                                 &      \cellcolor{rk3}0.796&       0.255& 22.63                                 &      0.821& \multicolumn{1}{l|}{0.231}      & 23.19                                 &      0.830&       0.214& 32.25&      0.922&  \multicolumn{1}{l|}{0.196}     & 34.43&       0.939&       0.169\\
% 2DGS*   & 32.89                                 &      \cellcolor{rk2}0.965& \multicolumn{1}{l|}{\cellcolor{rk1}0.039}      & 33.13                                 &      \cellcolor{rk3}0.968&       \cellcolor{rk2}0.031&  26.27 &      \cellcolor{rk2}0.765& \multicolumn{1}{l|}{\cellcolor{rk3}0.301}      & \cellcolor{rk2}27.01 &      \cellcolor{rk1}0.816&       \cellcolor{rk1}0.208& 22.88                                 &      0.824& \multicolumn{1}{l|}{0.223}      & 23.07                                 &      \cellcolor{rk3}0.836&       \cellcolor{rk2}0.189& 32.74                    &       \cellcolor{rk3}0.932&    \multicolumn{1}{l|}{ \cellcolor{rk3}0.181}   & 34.69                   &       \cellcolor{rk2}0.950&        \cellcolor{rk2}0.131\\
2DGS w/o. $\mathcal{L}_N$   & \cellcolor{rk3}33.08                              &\cellcolor{rk2}0.965     & \multicolumn{1}{l|}{\cellcolor{rk2}0.040}      & \cellcolor{rk3}33.65                                & \cellcolor{rk2}0.969    & \cellcolor{rk3}0.034      &26.28   & \cellcolor{rk3}0.760     & \multicolumn{1}{l|}{\cellcolor{rk3}0.302}      &\cellcolor{rk2}27.20  & \cellcolor{rk2}0.802   &\cellcolor{rk3}0.240       &    22.78                              & 0.823    & \multicolumn{1}{l|}{\cellcolor{rk3}0.219}      &  \cellcolor{rk3}23.40                                  &\cellcolor{rk3}0.836     & \cellcolor{rk3}0.201      & \cellcolor{rk3}33.88                  & \cellcolor{rk2}0.933     &    \multicolumn{1}{l|}{\cellcolor{rk3}0.188 }   & \cellcolor{rk3}36.35                 & \cellcolor{rk3}0.947     &  \cellcolor{rk3}0.165      \\
% 2DGS* w/o.$\mathcal{L}_N$   &                                  &      & \multicolumn{1}{l|}{}      &                                 &      &       & 26.68 & 0.772  & \multicolumn{1}{l|}{0.286}      &  &     &      &  23.17                              &0.831  & \multicolumn{1}{l|}{0.208}      &                                 &      &       &  35.12                  &0.940   &    \multicolumn{1}{l|}{0.177}   &                   &       &        \\
Ours-BI & \cellcolor{rk3}33.08 &      \cellcolor{rk2}0.965& \multicolumn{1}{l|}{\cellcolor{rk2}0.040}      & \cellcolor{rk2}33.69 &      \cellcolor{rk2}0.969&       \cellcolor{rk2}0.031& \cellcolor{rk3}26.31&      0.756& \multicolumn{1}{l|}{\cellcolor{rk2}0.300}      & 26.75                                 &      \cellcolor{rk3}0.796&       \cellcolor{rk2}0.233& \cellcolor{rk1}23.49 &      \cellcolor{rk1}0.833& \multicolumn{1}{l|}{\cellcolor{rk2}0.211}      & \cellcolor{rk1}23.75 &      \cellcolor{rk2}0.837&       \cellcolor{rk2}0.197& \cellcolor{rk1}34.34                    &       \cellcolor{rk2}0.933&    \multicolumn{1}{l|}{ \cellcolor{rk2}0.176}   &  \cellcolor{rk2}37.11                    &       \cellcolor{rk2}0.950&        \cellcolor{rk2}0.137\\
Ours-NN & \cellcolor{rk1}33.15 &      \cellcolor{rk1}0.966& \multicolumn{1}{l|}{\cellcolor{rk1}0.039}      & 33.39 &      \cellcolor{rk3}0.968&       0.036& \cellcolor{rk2}26.44  &      \cellcolor{rk3}0.762& \multicolumn{1}{l|}{0.305}      & \cellcolor{rk3}26.92 &      0.795&       0.258& \cellcolor{rk3}23.17&      \cellcolor{rk3}0.829& \multicolumn{1}{l|}{0.222}      & 23.26 &      0.830&       0.216&  \cellcolor{rk3}33.88                    &      \cellcolor{rk3}0.929&   \multicolumn{1}{l|}{0.201}    &  34.70                     &      0.935&       0.187\\
Ours-MK & \cellcolor{rk2}33.09 &      \cellcolor{rk2}0.965& \multicolumn{1}{l|}{\cellcolor{rk2}0.040}      & \cellcolor{rk1}34.10 &      \cellcolor{rk1}0.970&       \cellcolor{rk1}0.030& \cellcolor{rk1}26.55  &      \cellcolor{rk1}0.767& \multicolumn{1}{l|}{\cellcolor{rk1}0.293}      & \cellcolor{rk1}27.31 &      \cellcolor{rk2}0.815&       \cellcolor{rk2}0.209& \cellcolor{rk2}23.28 &      \cellcolor{rk2}0.831& \multicolumn{1}{l|}{\cellcolor{rk1}0.209}      & \cellcolor{rk2}23.72 &      \cellcolor{rk1}0.847&       \cellcolor{rk1}0.179&  \cellcolor{rk2}34.29                    &       \cellcolor{rk1}0.935&   \multicolumn{1}{l|}{ \cellcolor{rk1}0.174}    &  \cellcolor{rk1}37.76                    &       \cellcolor{rk1}0.957&        \cellcolor{rk1}0.117\\ \bottomrule
\end{tabular}
}
\vspace{-2mm}

    \label{tab:limitNum}
\end{table*}

\begin{table}[!tp]
    \centering
    \caption{Comparison on the Synthetic Blender~\cite{mildenhall2021nerf} dataset: 2DGS* is adjusted to match the total number of parameters used by our movable kernel method by increasing the number of Gaussian primitives. The notation w/o. \(\mathcal{L}_N\) indicates that no normal loss is applied. }
    \vspace{-2mm}
    \resizebox{0.9\linewidth}{!}{
\begin{tabular}{l|ccccc}
\toprule
      & 2DGS  & 2DGS* & 2DGS w/o. $\mathcal{L}_N$ & 2DGS* w/o. $\mathcal{L}_N$ &  Ours-MK \\ \midrule
PSNR  & 32.87 & 33.13 & \cellcolor{rk3}33.65    & \cellcolor{rk2}33.99     & \cellcolor{rk1}34.10   \\
SSIM  & 0.965 & \cellcolor{rk3}0.968 & \cellcolor{rk2}0.969    & \cellcolor{rk1}0.970     & \cellcolor{rk1}0.970   \\
LPIPS & 0.038 & \cellcolor{rk2}0.031 & \cellcolor{rk3}0.034    & \cellcolor{rk2}0.031     & \cellcolor{rk1}0.030   \\ \bottomrule
\end{tabular}
}
\vspace{-2mm}

    \label{tab:2dgs*}
\end{table}

\begin{table}[!tp]
    \centering
    \caption{PSNR on Synthetic Blender~\cite{mildenhall2021nerf}, Mip-NeRF360~\cite{barron2022mipnerf360} and Tanks\&Temples~\cite{knapitsch2017tanks} datasets. Note that all data except 2DGS~\cite{2dgs} are from the 3DGS paper. }
    \vspace{-2mm}
    \resizebox{\linewidth}{!}{
\begin{tabular}{l|ccc|ccc|ccc}
\toprule
            & \multicolumn{3}{c|}{Synthetic Blender}                       & \multicolumn{3}{c|}{Mip-NeRF360}                     & \multicolumn{3}{c}{Tanks\&Temples}                   \\ \midrule
Metrics      & \multicolumn{1}{l}{PSNR}              & SSIM & LPIPS & \multicolumn{1}{l}{PSNR}              & SSIM & LPIPS & \multicolumn{1}{l}{PSNR}              & SSIM & LPIPS \\ \midrule
Plenoxels~\cite{fridovich2022plenoxels}   & 31.76                                 &      -&       -& 23.08                                 &      0.626&       0.463& 21.08&      0.719&       0.379\\
INGP-Base~\cite{muller2022instant}   & 33.18 &      -&       -& 25.30                                 &      0.671&       0.371& 21.72                                 &      0.723&       0.330\\
Mip-NeRF360~\cite{barron2022mip} & 33.09                                 &      -&       -& 27.69 &      0.792&       0.237& 22.22                                 &      0.759&       0.257\\
\XR{Zip-NeRF~\cite{barron2023zip}} &33.10 & 0.971 & 0.031 & 28.54 & 0.828 & 0.189 & - & - &- \\
3DGS~\cite{3dgs}        & 33.32 &      -&       -& 27.21 &      0.815&       0.214& 23.14 &      0.841&       0.183\\
\XR{MCMC-3DGS~\cite{kheradmand20243d}} & 33.80                                 &      0.970&       0.040& 29.89                                &      0.900&       0.190& 24.29 &      0.860&       0.190\\
\XR{Mip-Splatting~\cite{yu2024mip}} &33.36 & 0.969 & 0.031 & 27.79 & 0.827 & 0.203 & - & - &- \\
\XR{Textured-GS~\cite{huang2024textured}} & 28.44      &    0.909 &      0.118 & 27.64                               &      0.825&       0.209& 23.49 &      0.843&       0.191\\
\XR{Textured-Gaus~\cite{chao2024textured}} & 33.17                                 &      0.965&       0.042& 27.37                                &      0.795&       0.187& 24.22 &      0.825&       0.172
\\ \midrule

2DGS~\cite{2dgs}        & 32.87                                 &      0.965&       0.038& 26.86                                 &      0.796&       0.255& 23.19 &      0.830&       0.214\\
\XR{PGSR~\cite{chen2024pgsr}} & 30.22                                 & 0.954     &       0.057& 27.25                                &      0.833&       0.178& 22.92 &0.857      &0.147       \\
Ours-MK     & 34.10 &      0.970&     0.030& 27.31 &      0.815&      0.209& 23.72 &      0.847&       0.179\\ \bottomrule
\end{tabular}
}
\vspace{-2mm}

    \label{tab:psnrOnDataSet}
\end{table}

% \begin{table}[!tp]
%     \centering
%     \resizebox{0.6\linewidth}{!}{
% \begin{tabular}{l|ccc}
% \toprule
%            & PSNR$\uparrow $  & SSIM$\uparrow$ & LPIPS$\downarrow$   \\ \midrule
% 2DGS     & \cellcolor{rk2}34.43& \cellcolor{rk2}0.939& \cellcolor{rk2}0.169\\
% Ours-MK      & \cellcolor{rk1}37.76                    &\cellcolor{rk1}0.957&\cellcolor{rk1}0.117\\ \bottomrule
% \end{tabular}
% }
% \vspace{-2mm}
% \caption{Novel View Synthesis comparison with 2DGS~\cite{2dgs} on the DTU~\cite{jensen2014DTU} dataset.}
%     \label{tab:dtu}
% \end{table}

\subsection{Implementation}
\name~is implemented based on the 2DGS~\cite{2dgs} code framework. 
We modified their CUDA kernels to implement our method and derived the corresponding back-propagation gradient code for each different spatially varying function. 
We followed all the setting parameters of 2DGS~\cite{2dgs} and 3DGS~\cite{3dgs} and compared them under the same conditions. 
As used in 2DGS~\cite{2dgs} and 3DGS~\cite{3dgs}, we trained for 30K iterations while keeping the gradient splitting threshold at 0.0002, resetting the opacity to 0.01 every 3000 iterations, and stopping the splitting, cloning and removing of Gaussians after 15K iterations. 
We discard the normal consistency loss because we only focus on the quality of novel view synthesis.
% , we only use the distortion loss of 2DGS~\cite{2dgs} to avoid occlusion caused by incorrect depth estimation.
% 2DGS~\cite{2dgs} adds depth distortion loss and normal consistency loss based on the loss function of 3DGS~\cite{3dgs}. Since we only focus on the quality of novel view synthesis, we only use the distortion loss of 2DGS~\cite{2dgs} to avoid occlusion caused by incorrect depth estimation. 
All our experiments were run on a single NVIDIA A100 80GB GPU and an Intel(R) Xeon(R) Platinum 8375C CPU.
% \name is implemented based on the 2DGS~\cite{2dgs} code framework, which is modified on the 3DGS~\cite{3dgs} code base. 
% We modified their CUDA kernels to implement our approach. 
% We followed all the setting parameters of 2DGS~\cite{2dgs} and 3DGS and compared them under the same parameters. 
% For each scene, we train it for 30K iterations while keeping the gradient splitting threshold at 0.0002, resetting the opacity to 0.01 every 3000 iterations, and not splitting and removing Gaussians after 15K iterations.
% 2DGS~\cite{2dgs} adds depth distortion loss and normal consistency loss based on the loss function of 3DGS~\cite{3dgs}. Since we only focus on the quality of novel view synthesis, we only use the distortion loss of 2DGS~\cite{2dgs} to avoid occlusion caused by some incorrect depth estimation.
% All our experiments were run on a single NVIDIA A100 80GB GPU and Intel(R) Xeon(R) Platinum 8375C CPU. 

\textbf{Dataset.}
Following 2DGS~\cite{2dgs} and 3DGS~\cite{3dgs}, we tested the Synthetic Blender dataset~\cite{mildenhall2021nerf} and Tanks\&Temples~\cite{knapitsch2017tanks} at their native resolution. We tested the DTU~\cite{jensen2014DTU} dataset at a resolution of $0.8K \times 0.6K$, which is one quarter of the native resolution. For the Mip-NeRF360~\cite{barron2022mipnerf360} dataset, we followed the 2DGS~\cite{2dgs} test settings, using the ``images\_4'' setting for outdoor scenes, which has a resolution of about $0.8K \times 0.6K$, and the ``images\_2'' setting for indoor scenes, which has a resolution of about $1.0K \times 0.8K$. The pictures and quantitative data (except DTU~\cite{jensen2014DTU}) we show are calculated and rendered on the test set, which never appeared in the training set.

\begin{figure}
    \centering
    \vspace{-2mm}
    \begin{overpic}[width=0.98\linewidth]{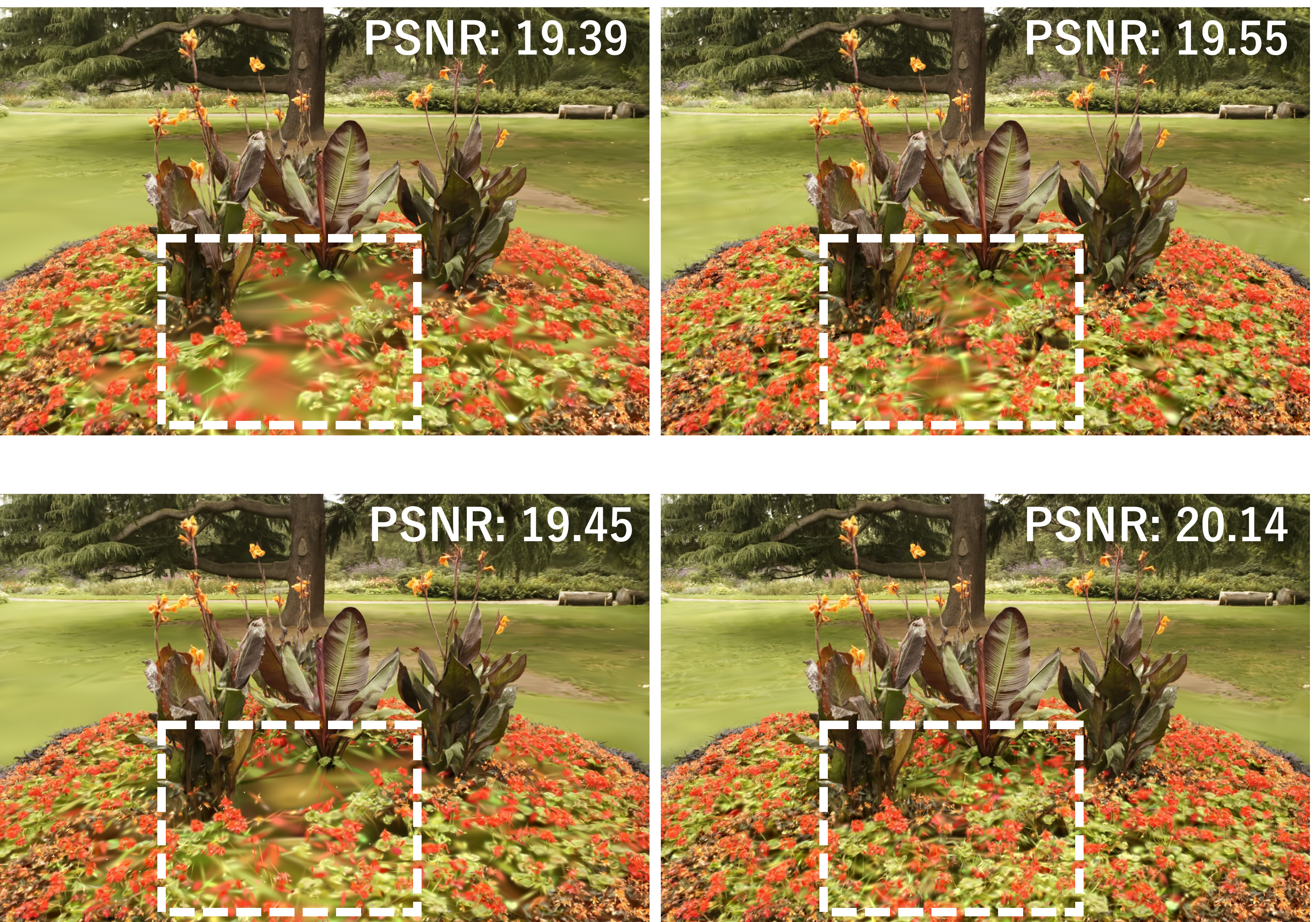}
    \put(18, 33.5){2DGS}
    \put(67, 33.5){Ours-BI}
    \put(16, -3.5){Ours-NN}
    \put(66, -3.5){Ours-MK}
    \end{overpic}
    \vspace{0pt}
    \caption{
\XR{Visual comparison between three different spatially varying functions and 2DGS~\cite{2dgs}.}
    }
\label{fig:difffunc}
\vspace{-2mm}
\end{figure}

\begin{figure}[!tp]
    \centering
    \begin{overpic}[width=0.98\linewidth]{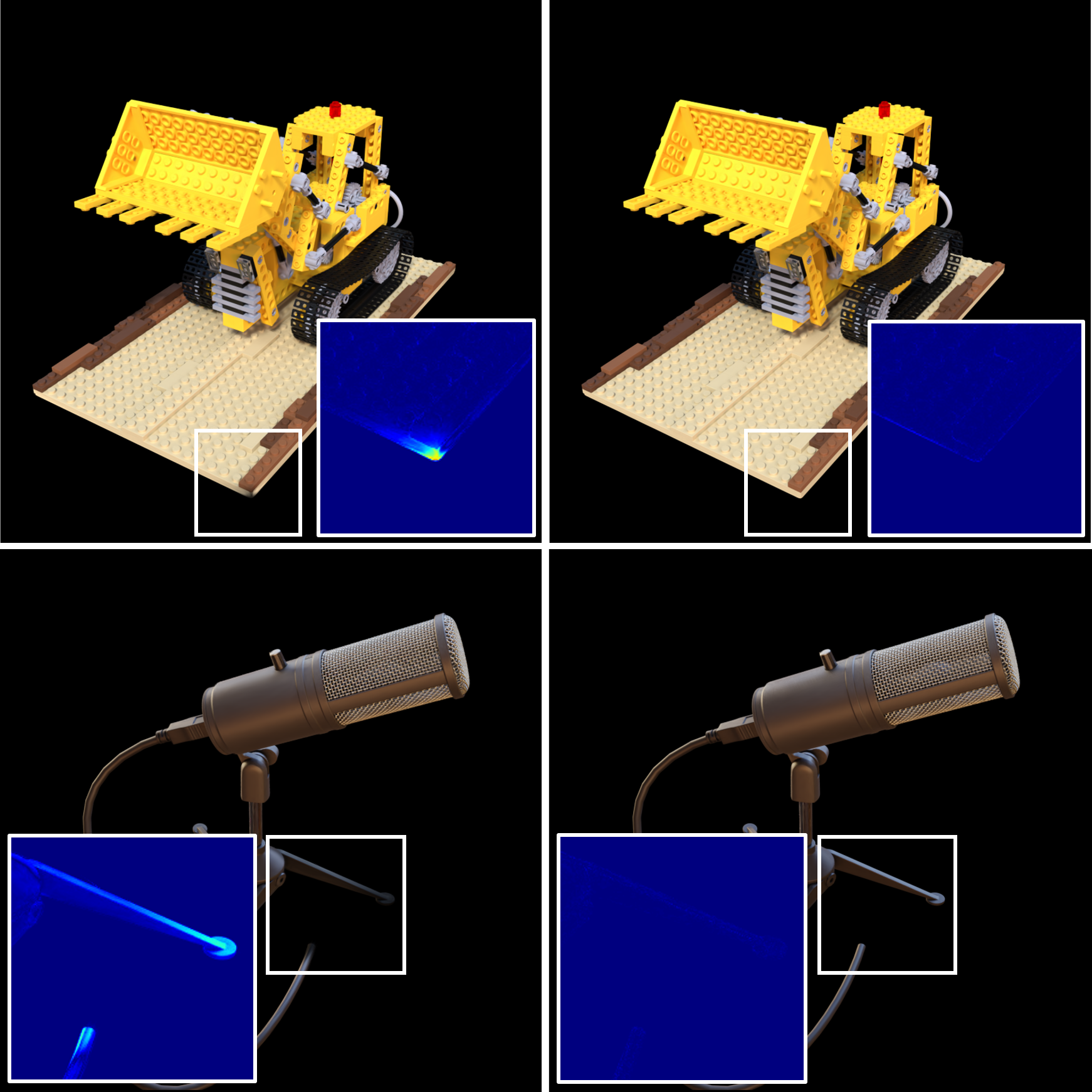}
    % \put(18, 33.5){2DGS}
    % \put(67, 33.5){Ours-BI}
    \put(20, -3.5){2DGS}
    \put(66, -3.5){Ours-MK}
    \end{overpic}
    \vspace{0pt}
    \caption{Visualization comparison with 2DGS~\cite{2dgs} on the Synthetic Blender~\cite{mildenhall2021nerf} dataset. The blue zoom-in windows show the error map against the ground truth image.
    }
\label{fig:2dgscomp}
\vspace{-4mm}
\end{figure}

\begin{figure*}[!tp]
    \centering
    \begin{overpic}[width=\linewidth]{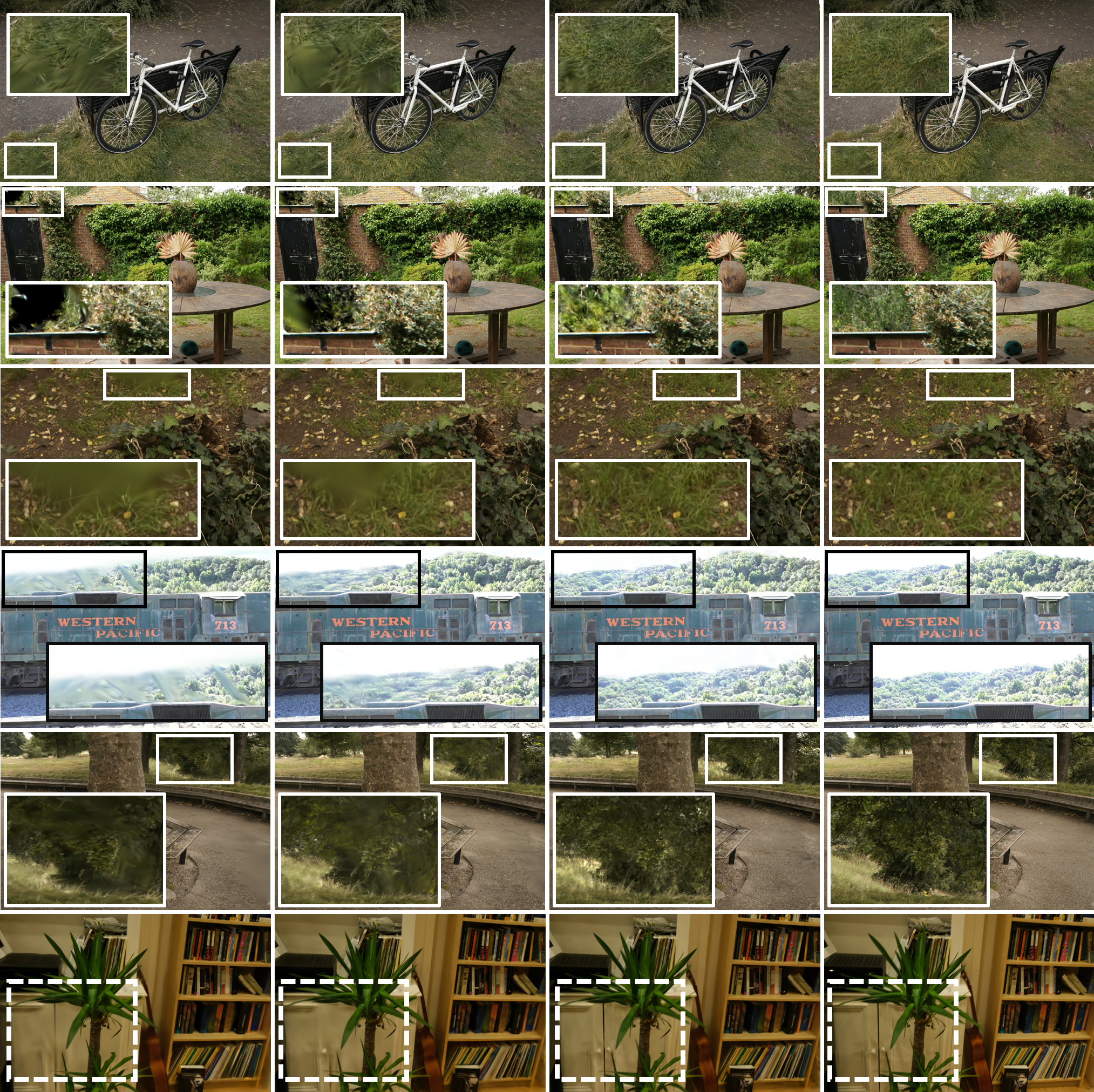}
    \put(9,  -1.7){2DGS}
    \put(34, -1.7){3DGS}
    \put(58, -1.7){Ours-MK}
    \put(82, -1.7){Ground Truth}
    \end{overpic}
    \vspace{-10pt}
    \caption{
Visual comparison with 2DGS~\cite{2dgs} and 3DGS~\cite{3dgs} on both the Mip-NeRF360~\cite{barron2022mipnerf360} dataset and the Tanks\&Temples~\cite{knapitsch2017tanks} dataset shows that \name~can reconstruct details better due to stronger expressiveness. 
    }
\label{fig:comp}
% \vspace{-4mm}
\end{figure*}

\subsection{Comparison}
\label{sec:comp}

\textbf{Dataset.}
% We tested our method on multiple datasets including Synthetic Blender~\cite{mildenhall2021nerf}, DTU~\cite{jensen2014DTU}, Mip-NeRF360~\cite{barron2022mipnerf360} and Tanks\&Temples~\cite{knapitsch2017tanks} datasets, and we referred to the resolution settings of 2DGS and 3DGS.
% We test the Synthetic Blender dataset~\cite{mildenhall2021nerf} with native resolution and the DTU~\cite{jensen2014DTU} dataset with $0.8K \times 0.6K$ resolution, which is one quarter of the native resolution.
% For the Mip-NeRF360~\cite{barron2022mipnerf360} dataset, we follow the 2DGS test settings, using the ``images\_4'' setting for outdoor scenes, which has a resolution of about $0.8K \times 0.6K$. For indoor scenes, we use the ``images\_2'' setting, which has a resolution of about $1.0K \times 0.8K$. 
We tested our method on multiple datasets, including Synthetic Blender~\cite{mildenhall2021nerf}, DTU~\cite{jensen2014DTU}, Mip-NeRF360~\cite{barron2022mipnerf360}, and Tanks\&Temples~\cite{knapitsch2017tanks}, and we follow the same evaluation settings of 2DGS~\cite{2dgs} and 3DGS~\cite{3dgs} including the image resolution and the choice test set.
% To measure the difference between the reconstructed image and the ground truth, we use three indicators: \textit{Peak Signal-to-Noise Ratio} (PSNR), \textit{Structural Similarity} (SSIM), and \textit{Learned Perceptual Image Patch Similarity} (LPIPS). 
% For the geometric reconstruction in the DTU~\cite{jensen2014DTU} dataset, we use the \textit{Chamfer Distance} (CD) to measure the difference between the reconstructed mesh and the ground truth mesh.
We use PSNR, SSIM~\cite{brunet2011mathematical}, and LPIPS~\cite{zhang2018perceptual} to measure the performance on all datasets for the novel-view-synthesis task. For the surface reconstruction, use Chamfer Distance (CD) to measure the accuracy of geometry on the DTU~\cite{jensen2014DTU} dataset.

% We use three different colors to highlight the top three in the subsequent tables, with the darkest red representing first place and the lightest yellow representing third place.

\textbf{Comparison on Different Spatially Varying Functions.}
First, we present the comparison between 2DGS~\cite{2dgs} and three of our spatially varying functions in Table~\ref{tab:limitNum}. 
% For per-scene results, please refer to the supplementary material.
It can be seen that on most datasets, our movable kernel achieves the best reconstruction quality for the novel-view-synthesis task. The spatially varying function of bilinear interpolation also achieves good results in some of the scenes. The tiny neural network performs well when the number of Gaussian primitive is limited, demonstrating its strong representation ability. However, optimizing a neural network is usually difficult with unstable convergence, thus performing worse than the other two functions without limiting primitive numbers.
% For more detailed data, please refer to Table~\ref{tab:3settingOnNerf}, Table~\ref{tab:3settingOn360}, and Table~\ref{tab:3settingOnTaTDB} in Appendix~\ref{sec:A_tables}.
To avoid errors caused by different numbers of Gaussian points and to further demonstrate our stronger expressiveness, we also tested the original 2DGS~\cite{2dgs} and 2DGS without normal loss in a limited number of Gaussians, as shown in Table~\ref{tab:limitNum}. 

To ensure fair comparisons, we scale the number of Gaussians in standard 2DGS proportionally to match our method's total parameter count in Table~\ref{tab:2dgs*}, creating 2DGS*. For instance, when our approach uses 10,000 Gaussians (40\% more parameters per primitive), 2DGS* is adjusted to 14,000 primitives, maintaining parameter parity while preserving 2DGS's original structure to isolate architectural improvements from parameter scaling.
While our primitive design increases per-element parameters by 40\% compared to 2DGS, the system achieves superior rendering quality with fewer total primitives and reduced overall parameters. This efficiency stems from enhanced representational capacity per primitive, enabling more compact scene encoding.
% For a fair comparison, we provide 2DGS*, which uses the same total number of parameters as our movable kernel method by increasing the number of Gaussian primitives, and they are not following the number limit in the table. 

% However, it cannot achieve optimal results without Gaussian number limit due to its difficulty in converging.

Additionally, the spatially varying functions based on bilinear interpolation and tiny neural networks surpassed 2DGS~\cite{2dgs}, proving the effectiveness of these spatially varying functions. Fig.~\ref{fig:difffunc} shows the visual comparison of these three spatially varying functions. It can be seen that although the movable kernel function does not have the largest number of parameters (as shown in Fig.~\ref{fig:paras}), it has the best ability to reconstruct details. 
While the other two spatially varying functions cannot reconstruct all details as well as the movable kernel, they still show significant improvements compared to 2DGS~\cite{2dgs}. 
% We also provide a detailed comparison with 2DGS in Appendix A and more discussions in Appendix D.

\begin{figure}[!tp]
    \centering
    \begin{overpic}[width=0.98\linewidth]{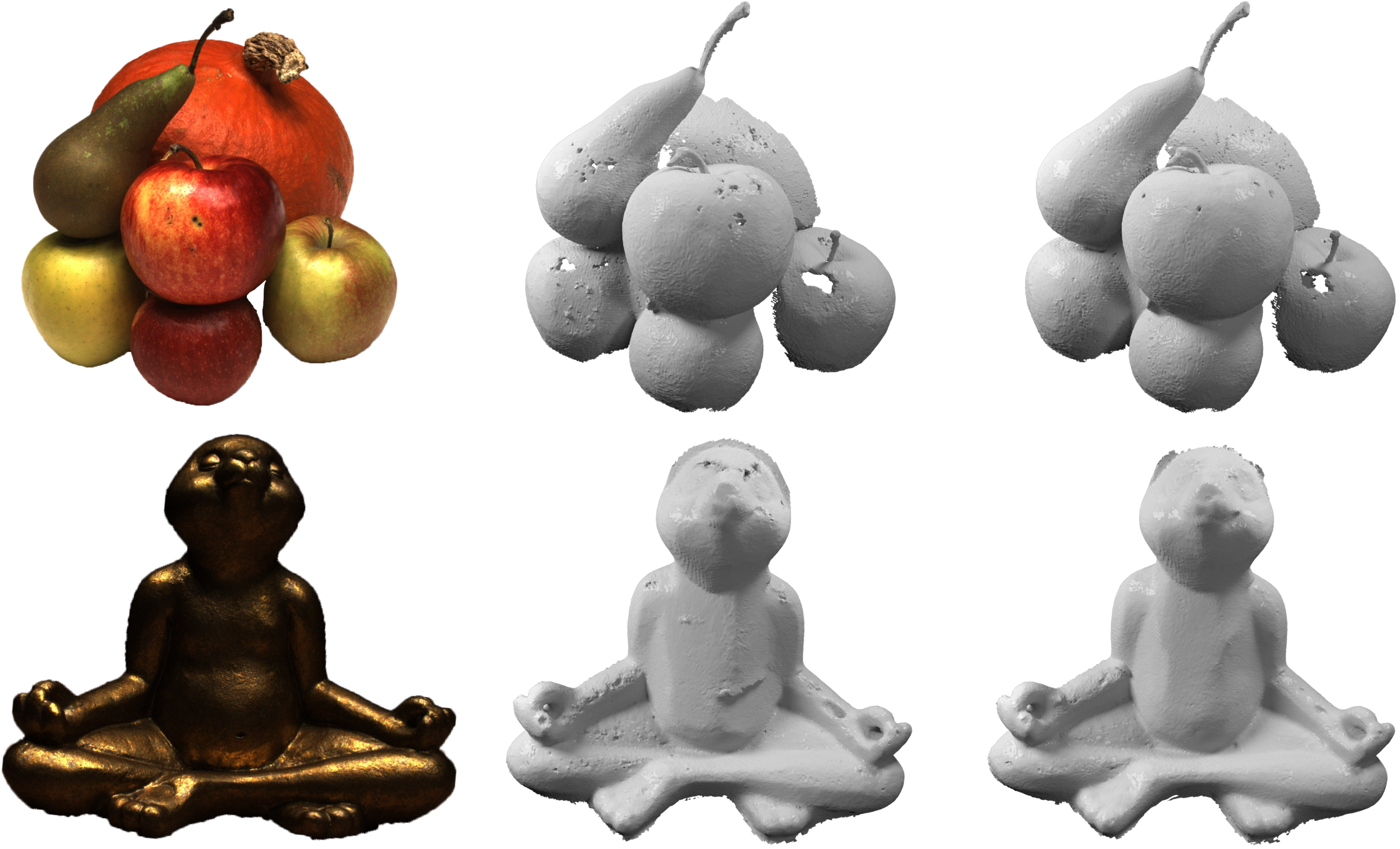}
    % \put(18, 33.5){2DGS}
    % \put(67, 33.5){Ours-BI}
    \put(10, -3.5){Image}
    \put(46, -3.5){2DGS}
    \put(77, -3.5){Ours-MK}
    \end{overpic}
    \vspace{1pt}
    \caption{
Visual comparison between three different spatially varying functions and 2DGS~\cite{2dgs}.
    }
\label{fig:Recon}
\vspace{-4mm}
\end{figure}

\textbf{Novel View Synthesis.}
We compare the best setting of \name~with other state-of-the-art (SoTA) methods. Table~\ref{tab:psnrOnDataSet} shows the PSNR comparison across different datasets. It can be observed that we have achieved optimal or near-optimal results on multiple datasets. 
\XR{
% Although we do not outperform all other methods on Mip-NeRF360~\cite{barron2022mipnerf360} dataset (note they use more representative 3D representation or implicit representation but we use 2D explicit representation), we still surpass some explicit 2D Gaussian-based reconstruction methods.
Our method substantially enhances novel-view synthesis capability while maintaining high-quality geometric reconstruction. 
In particular, our approach excels in scenes characterized by relatively simple geometry but complex appearance or textures. 
This strength is clearly reflected in the results on the Blender~\cite{mildenhall2021nerf} dataset, where our method surpasses all state-of-the-art techniques, regardless of whether they are based on 2DGS~\cite{2dgs}, 3DGS~\cite{3dgs}, or NeRF~\cite{mildenhall2021nerf}.
Objects in the Blender~\cite{mildenhall2021nerf} dataset typically exhibit smooth surfaces and limited geometric complexity, yet rich and detailed textures. 
Under such conditions, SVGS effectively captures intricate texture variations and high-frequency appearance details without sacrificing rendering consistency, demonstrating its strong expressiveness and robustness in texture-dominated scenarios.
While our results on the Mip-NeRF360~\cite{barron2022mipnerf360} dataset are slightly behind those of methods employing more expressive 3D or implicit representations~\cite{barron2023zip, kheradmand20243d} (SVGS use explicit 2D representation), SVGS still outperforms all explicit 2D Gaussian-based reconstruction approaches.
It is worth noting that although our method performs slightly worse than 3DGS-based approaches such as MCMC-3DGS~\cite{kheradmand20243d} and Textured-Gaus~\cite{chao2024textured} on the Mip-360 dataset~\cite{barron2022mip}, this gap primarily stems from the inherently stronger expressive power of 3DGS compared to 2DGS. Nonetheless, our method achieves comparable results under the 2DGS framework. Moreover, our approach and MCMC-3DGS~\cite{kheradmand20243d} address Gaussian splatting from two distinct yet complementary perspectives. In theory, they are not mutually exclusive and could be integrated, which presents a promising direction for future research.
This distinction also reflects our overarching design goal:
rather than pursuing purely higher PSNR, we seek a balanced trade-off between geometry reconstruction fidelity and novel-view synthesis (NVS) accuracy. Under this principle, SVGS maintains high geometric reconstruction quality while significantly improving NVS performance. It is worth emphasizing that methods achieving higher NVS precision often exhibit inferior reconstruction quality (like 3DGS~\cite{3dgs} and  Nerf~\cite{mildenhall2021nerf}), whereas those with superior geometric fidelity tend to underperform in NVS (NeuS~\cite{wang2021neus} and PGSR~\cite{chen2024pgsr}). SVGS thus occupies a balanced middle ground, delivering strong results in both aspects.
}

\begin{table}[!tp]
    \centering
    \caption{\XR{Chamfer distance and PSNR comparison of geometric reconstruction on the DTU~\cite{jensen2014DTU} dataset under different number constraints.}}
    \vspace{-3mm}
    \resizebox{0.95\linewidth}{!}{
\begin{tabular}{l|cc|cc|cc}
\toprule
Nums & \multicolumn{2}{c|}{50K} & \multicolumn{2}{c|}{100K} & \multicolumn{2}{c}{noLimit} \\ \midrule
Metrics  & CD $\downarrow$         & PSNR $\uparrow$      & CD $\downarrow$         & PSNR $\uparrow$       & CD $\downarrow$          & PSNR $\uparrow$   \\ \midrule

2DGS & 1.11        & \cellcolor{rk3}28.76 & \cellcolor{rk3}0.84 & \cellcolor{rk3}32.25 & \cellcolor{rk3}0.75 & \cellcolor{rk3}34.43 \\
2DGS* & \cellcolor{rk3}1.00 & 28.26 & \cellcolor{rk3}0.84 & \cellcolor{rk2}32.74 & \cellcolor{rk2}0.74 & \cellcolor{rk2}34.69 \\
\XR{PGSR} & \cellcolor{rk1}0.73 & \cellcolor{rk2}29.83 & \cellcolor{rk1}0.60 & 31.32 & \cellcolor{rk1}0.52 & 33.38 \\
Ours-MK & \cellcolor{rk2}0.99 & \cellcolor{rk1}29.86 & \cellcolor{rk2}0.82 & \cellcolor{rk1}32.80 & 0.76 & \cellcolor{rk1}35.22 \\ 
\bottomrule
\end{tabular}
}
\vspace{-2mm}
    \label{tab:GeoRecon}
\end{table}

Fig.~\ref{fig:2dgscomp} presents a visual comparison between our results and those of 2DGS~\cite{2dgs} on the Synthetic Blender~\cite{mildenhall2021nerf} dataset. We provide the error map alongside the ground truth for easier observation. It is evident that 2DGS~\cite{2dgs} struggles with reconstructing shape boundaries and abrupt parts. For instance, in the ``mic'' model, its wires and brackets are not clearly visible from certain angles, whereas our method addresses these issues effectively.

Fig.~\ref{fig:comp} demonstrates the visual comparison with 2DGS~\cite{2dgs} and 3DGS~\cite{3dgs}, showcasing the powerful expressiveness of \name. For example, in the first row, the \textit{bicycle} example highlights our ability to capture finer details. The distant view of the \textit{garden} example in the second row and the \textit{train} in the fourth row emphasize our method's expressiveness of corner cases that appear less common in the training views. 
Additionally, since our method can express more details, we can reduce the smearing effect, as demonstrated by the \textit{leaves} in the fifth row.
\FF{Nevertheless, since \name \space does not explicitly include an anti-aliasing mechanism, it may still struggle to fully recover very fine high-frequency details in some challenging regions. For instance, the distant bushes of the \textit{garden} example, some blur remains compared to the ground truth.}

Table~\ref{tab:psnrOnDataSet} also shows the comparison results between our method and 2DGS~\cite{2dgs} on the DTU~\cite{jensen2014DTU} dataset. It should be noted that, following 2DGS~\cite{2dgs}, the DTU~\cite{jensen2014DTU} dataset does not include a test set, so we only report the metrics on the training set. It can be seen that our method significantly surpasses 2DGS~\cite{2dgs}.
\XR{
Detailed versions of the corresponding tables can be founded in our supplementary materials.}

\begin{table}[!tp]
    \centering
    \caption{\XR{Comparison with more state-of-the-art (SOTA) methods on the DTU~\cite{jensen2014DTU} dataset without number limit.}}
    \vspace{-3mm}
    \resizebox{0.98\linewidth}{!}{
\begin{tabular}{l|cccccccc}
\toprule
Metric & 2DGS & 2DGS* & \XR{PGSR} & NeuS & 3DGS & \XR{VolSDF} & SuGaR & Ours-MK \\ \midrule
CD $\downarrow$ &
\cellcolor{rk3}0.75 &
\cellcolor{rk2}0.74 &
\cellcolor{rk1}0.52 &
0.84 &
1.96 &
0.86 &
1.33 &
\cellcolor{rk4}0.76 \\ 
PSNR $\uparrow$ &
34.43 &
\cellcolor{rk3}34.69 &
33.38 &
31.97 &
\cellcolor{rk1}35.76 &
30.38 &
\cellcolor{rk4}34.57 &
\cellcolor{rk2}35.22 \\
\bottomrule
\end{tabular}
}
\vspace{-2mm}
\label{tab:GeoRecon_sota}
\end{table}

% \vspace{-4mm}
\textbf{Geometry Reconstruction.}
2DGS~\cite{2dgs} has demonstrated impressive capabilities in geometric reconstruction. Since \name~is implemented based on 2DGS~\cite{2dgs}, it inherits these excellent geometric reconstruction capabilities, although this is not our primary objective.
To evaluate our method, we enable the normal consistency loss from 2DGS~\cite{2dgs}, which is important for geometry reconstruction. We test our geometric reconstruction ability on the commonly used DTU dataset~\cite{jensen2014DTU} with different limits on the number of Gaussians, and report the quantitative analysis in Table~\ref{tab:GeoRecon}, Table~\ref{tab:GeoRecon_sota} and visualization results in Fig.~\ref{fig:Recon}.
The results demonstrate that our geometric reconstruction capabilities are comparable to 2DGS~\cite{2dgs} when there is no limit on the number of Gaussians, while also ensuring the highest PSNR, indicating the best image reconstruction quality. Furthermore, when a limited number of Gaussians is used, our powerful expressiveness becomes apparent, greatly outperforming 2DGS~\cite{2dgs} in both geometric reconstruction quality and image rendering quality.
\XR{As a recent method emphasizing geometric reconstruction, PGSR~\cite{chen2024pgsr} substantially surpasses 2DGS~\cite{2dgs} in reconstruction metrics; however, it does not achieve better rendering quality (PSNR), even when the number of Gaussians is restricted.}

We also present a detailed comparison of geometric reconstruction on the DTU~\cite{jensen2014DTU} dataset. 
% \sout{Table~\ref{tab:GeoReconA} compares our reconstruction results with other existing methods without Gaussian number restrictions.} 
The methods compared include implicit methods such as VolSDF~\cite{yariv2021volsdf}, and NeuS~\cite{wang2021neus}, as well as explicit methods such as 3DGS~\cite{3dgs} and SuGaR~\cite{guedon2024sugar}. 
We also provide the PSNR of image quality in the last column. 

\begin{figure*}[!htp]
    \centering
    \begin{overpic}[width=0.99\linewidth]{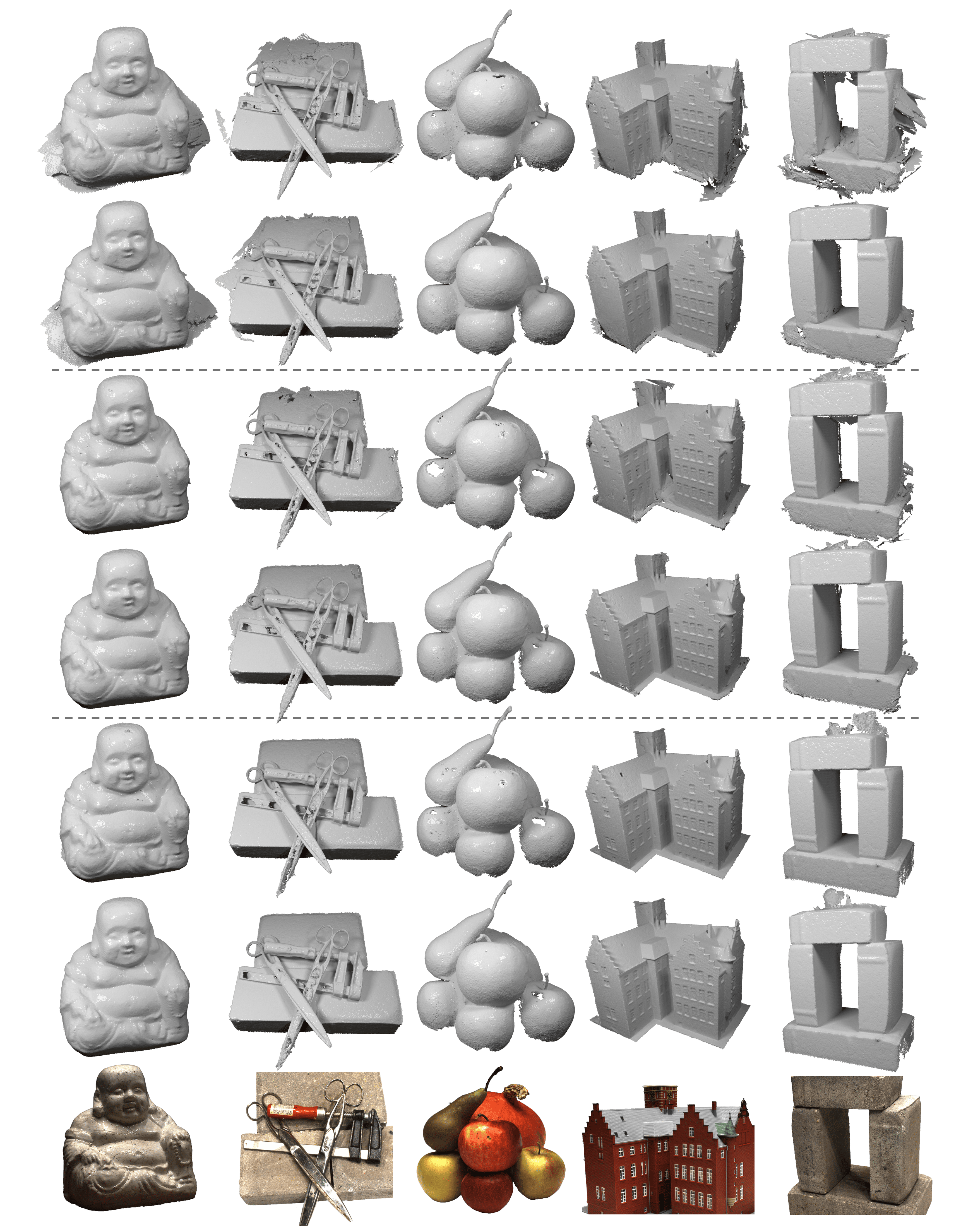}
    \put(1, 0) {\rotatebox{90}{\textbf{Ground Truth Image}}}
    \put(1, 15.5) {\rotatebox{90}{\textbf{Ours-NoLimit}}}
    \put(1, 29) {\rotatebox{90}{\textbf{2DGS-NoLimit}}}
    \put(1, 45){\rotatebox{90}{\textbf{Ours-100K}}}
    \put(1, 58){\rotatebox{90}{\textbf{2DGS-100K}}}
    \put(1, 73){\rotatebox{90}{\textbf{Ours-50K}}}
    \put(1, 87){\rotatebox{90}{\textbf{2DGS-50K}}}
    \end{overpic}
    \vspace{-6pt}
    \caption{
    More geometric reconstruction results on the DTU~\cite{jensen2014DTU} dataset. We show the comparison between 2DGS~\cite{2dgs} and our method under different Gaussian number limitations.
    }
\label{fig:moreRecon}
% \vspace{-4mm}
\end{figure*}

% \sout{Tables~\ref{tab:GeoRecon50k} and~\ref{tab:GeoRecon100k} further compare our geometric reconstruction with 2DGS when the number of Gaussians is limited to 50K and 100K. }
More visualization results are shown in Fig.~\ref{fig:moreRecon}.
It can be observed that \name~has a clear advantage over 2DGS under a limited number of Gaussians.

% 2DGS~\cite{2dgs} has demonstrated impressive capabilities in geometric reconstruction. Since \name is implemented based on 2DGS~\cite{2dgs}, it inherits these excellent geometric reconstruction capabilities, although this is not our primary objective.
% To evaluate our method, we enable the normal consistency loss from 2DGS~\cite{2dgs}, which is important for geometry reconstruction. We test our geometric reconstruction ability on the commonly used DTU dataset~\cite{jensen2014DTU} with different limits of Gaussian numbers, and report the quantitative analysis in Table~\ref{tab:GeoRecon} and visualization results in Fig.~\ref{fig:Recon}.
% The results demonstrate that our geometric reconstruction capabilities are comparable to those of 2DGS~\cite{2dgs} when their is no limit of Gaussian numbers, while also ensuring the highest PSNR, indicating the best image reconstruction quality.
% However, when a limited number of Gaussians is used, our powerful expression ability becomes apparent, greatly outperforming 2DGS~\cite{2dgs} in both geometric reconstruction quality and image rendering quality.

\begin{table}[!tp]
    \centering
    \caption{Ablation on the normal consistency loss, we present the comparative results of 2DGS~\cite{2dgs} and our method on the Synthetic Blender dataset~\cite{mildenhall2021nerf} dataset, both with and without normal loss.}
    \vspace{-2mm}
    \resizebox{0.7\linewidth}{!}{
\begin{tabular}{l|ccc}
\toprule
           & PSNR$\uparrow $  & SSIM$\uparrow$ & LPIPS$\downarrow$   \\ \midrule
2DGS-w.$\mathcal{L}_N$                 & 32.87& \cellcolor{rk3}0.965& 0.038\\
2DGS-w/o.$\mathcal{L}_N$            & \cellcolor{rk3}33.65& \cellcolor{rk2}0.969& \cellcolor{rk3}0.034\\
Ours-w.$\mathcal{L}_N$         & \cellcolor{rk2}33.85& \cellcolor{rk1}0.970& \cellcolor{rk2}0.031\\
Ours-w/o.$\mathcal{L}_N$       & \cellcolor{rk1}34.10& \cellcolor{rk1}0.970& \cellcolor{rk1}0.030\\ \bottomrule
\end{tabular}
}
\vspace{-2mm}

    \label{tab:normal}
\end{table}

\begin{table}[!tp]
    \centering
    \caption{Ablation of the number of kernels and kernel function forms on the Synthetic Blender dataset~\cite{mildenhall2021nerf}.}
    \resizebox{0.88\linewidth}{!}{
\begin{tabular}{l|cccc}
\toprule
           & PSNR$\uparrow $  & SSIM$\uparrow$ & LPIPS$\downarrow$ & Parameters$\downarrow$  \\ \midrule
% 2DGS~\cite{2dgs}     & ?& ?    & ?        \\
8 Kernels      & \cellcolor{rk2}34.04& \cellcolor{rk1}0.970& \cellcolor{rk1}0.030& \cellcolor{rk3}105      \\
Sigmoid Kernel       & \cellcolor{rk3}33.97& \cellcolor{rk2}0.969& \cellcolor{rk1}0.030& \cellcolor{rk1}81  \\
Ours-MK       & \cellcolor{rk1}34.10& \cellcolor{rk1}0.970& \cellcolor{rk1}0.030& \cellcolor{rk1}81  \\ \bottomrule
\end{tabular}
}
\vspace{-2mm}

    \label{tab:ablation}
\end{table}

\subsection{Ablation Study}
\label{sec:ablation}

\textbf{Normal Consistency Loss.}
Since our goal is to improve the novel view reconstruction capability of Gaussian surfels rather than their geometric reconstruction capability, we do not use the normal consistency loss provided by 2DGS~\cite{2dgs}. We report the ablation study of this loss in Table~\ref{tab:normal}, where we tested our method and the 2DGS~\cite{2dgs} method on the Synthetic Blender dataset~\cite{mildenhall2021nerf} both with and without normal consistency loss. 
All indicators showed that our method is significantly better than the baseline method, regardless of whether normal consistency loss is used.

\begin{table}[!tp]
    \centering
    \caption{\XR{Ablation on the number of layers in our tiny MLP network, evaluated on the \textit{Lego} scene of the Blender~\cite{mildenhall2021nerf} dataset.}}
    \resizebox{0.7\linewidth}{!}{
    \begin{tabular}{l|ccc}
    \toprule
    & PSNR$\uparrow$ & SSIM$\uparrow$ & LPIPS$\downarrow$ \\ \midrule
    NN-1Layer & 35.66 & 0.979 & 0.021 \\
    NN-2Layer & 35.44 & 0.979 & 0.021 \\
    NN-3Layer & 35.39 & 0.979 & 0.021 \\
    NN-4Layer & 35.38 & 0.979 & 0.021 \\
    Ours-MK   & \textbf{36.49} & \textbf{0.983} & \textbf{0.008} \\
    \bottomrule
    \end{tabular}
    }
    \vspace{-2mm}
    \label{tab:mlpablation}
\end{table}

\textbf{Design of Kernels.}
We conducted ablations on the design of the optimal form of movable kernels in our spatially varying function, as shown in Table~\ref{tab:ablation}. In all other experiments, we use 4 movable kernels, and we show here how the results change when using 8 kernels on the Synthetic Blender dataset~\cite{mildenhall2021nerf}. Similarly, we also tried changing the form of the kernel function from an exponential function $\mathcal{F}_{\mathbf{K}_i}$ to a sigmoid function $\mathcal{F}_{\mathbf{S}_i}$:
\begin{equation}
    \mathcal{F}_{\mathbf{S}_i}(\mathbf{p}) = 1 - \tanh(\left\| \mathbf{p} - \mathbf{K}_i \right\|^2)
\end{equation}
where \(\tanh(x) = \frac{\sinh(x)}{\cosh(x)}\). Table~\ref{tab:ablation} demonstrates the superiority of our method. Increasing the number of kernels to eight does not significantly improve our results and leads to an increase in the number of parameters. Replacing the kernel function can achieve results close to ours, but still cannot surpass our method.

\begin{table}[!tp]
    \centering
    \caption{Compared with 2DGS~\cite{2dgs}, which has more Gaussian points on the Synthetic Blender dataset~\cite{mildenhall2021nerf}, ``Grad'' indicates that the number of Gaussians in 2DGS~\cite{2dgs} is twice than \name~by modifying the gradient split threshold. ``Split'' means that the Gaussians is split into twice as many as \name~at the last split iteration of 2DGS. \dag indicates that the normal consistency loss is not used.
% For the number of parameters, we use the number of parameters used by our method as a baseline and show the relative number of parameters of 2DGS~\cite{2dgs} when using double and quadruple the number of Gaussians.
}
    \resizebox{0.8\linewidth}{!}{
\begin{tabular}{l|cccccc}
\toprule
           & PSNR$\uparrow$   & SSIM$\uparrow$ & LPIPS$\downarrow$  & GS Num$\downarrow$ \\ \midrule
2DGS-Grad    & 32.58 & \cellcolor{rk3}0.957    & 0.038     & 446K   \\
2DGS-Split    & 32.65 & \cellcolor{rk2}0.959    & 0.038     & 392K   \\
2DGS\dag-Grad  & \cellcolor{rk2}34.05 & \cellcolor{rk1}0.970       & \cellcolor{rk1}0.028   & \cellcolor{rk3}387K   \\
2DGS\dag-Split  & \cellcolor{rk3}33.94 & \cellcolor{rk1}0.970      &\cellcolor{rk2}0.029  & \cellcolor{rk2}384K \\
Ours-MK      & \cellcolor{rk1}34.10 & \cellcolor{rk1}0.970     & \cellcolor{rk3}0.030    & \cellcolor{rk1}205K \\ \bottomrule
\end{tabular}
}
\vspace{-2mm}

    \label{tab:4x2dgs}
\end{table}

\XR{
\textbf{Layers of MLP.}
We also investigated the effect of network capacity on our model through an ablation study of the tiny MLP module, varying the number of layers from 1 to 4. This experiment was designed to assess how network depth (and consequently, parameter count) influences reconstruction quality. Table~\ref{tab:mlpablation} reports quantitative results after 30K training iterations on the \textit{Lego} scene from the Blender~\cite{mildenhall2021nerf} dataset. The results show that increasing the number of layers has a negligible impact on reconstruction performance, with only a marginal decrease in PSNR. This finding reinforces our conclusion that larger networks introduce training instability without improving expressiveness. And it is far from our movable kernels version.}

\subsection{Comparison with 2DGS}
To facilitate a fair comparison with 2DGS~\cite{2dgs} and further demonstrate our enhancement of Gaussian expressiveness, we tested our method using the same or fewer parameters as 2DGS~\cite{2dgs} in Table~\ref{tab:4x2dgs}. As seen in Fig.~\ref{fig:paras}, for each Gaussian, the number of parameters used by \name~(MK) is about 1.4 times that of the original 2DGS~\cite{2dgs}. To avoid the misconception that our advantage lies solely in the number of parameters, we controlled the number of Gaussians in 2DGS~\cite{2dgs} to be twice that of \name~for comparison. When 2DGS~\cite{2dgs} uses twice the number of Gaussians, the number of parameters is approximately 1.43 times that of \name.
To preempt misinterpretation that our method merely splits Gaussians, 
we used two methods to increase the number of points in 2DGS~\cite{2dgs} in Table~\ref{tab:4x2dgs}. One method was directly modify the threshold of gradient to allow 2DGS~\cite{2dgs} to split and clone more freely (``Grad'' in the table). The other method was split 2DGS~\cite{2dgs} into twice the number of \name~at the final split iteration (``Split'' in the table). 
This modified 2DGS requires significantly more parameters than ours. Therefore, it is an unfair comparison for our approach but only to demonstrate our strong expressive power.
It can be seen that even when 2DGS~\cite{2dgs} uses twice the number of Gaussians as ours, which equates to 1.43 times the number of parameters, its expressiveness is still weaker than ours. This holds true regardless of whether the normal consistency loss is used, demonstrating the powerful scene representation capabilities of our spatially varying function.

\subsection{Discussion on Spatially Varying Functions}
\label{Asec:DissOnSVF}
\begin{figure}[!tp]
    \centering
    \begin{overpic}[width=\linewidth]{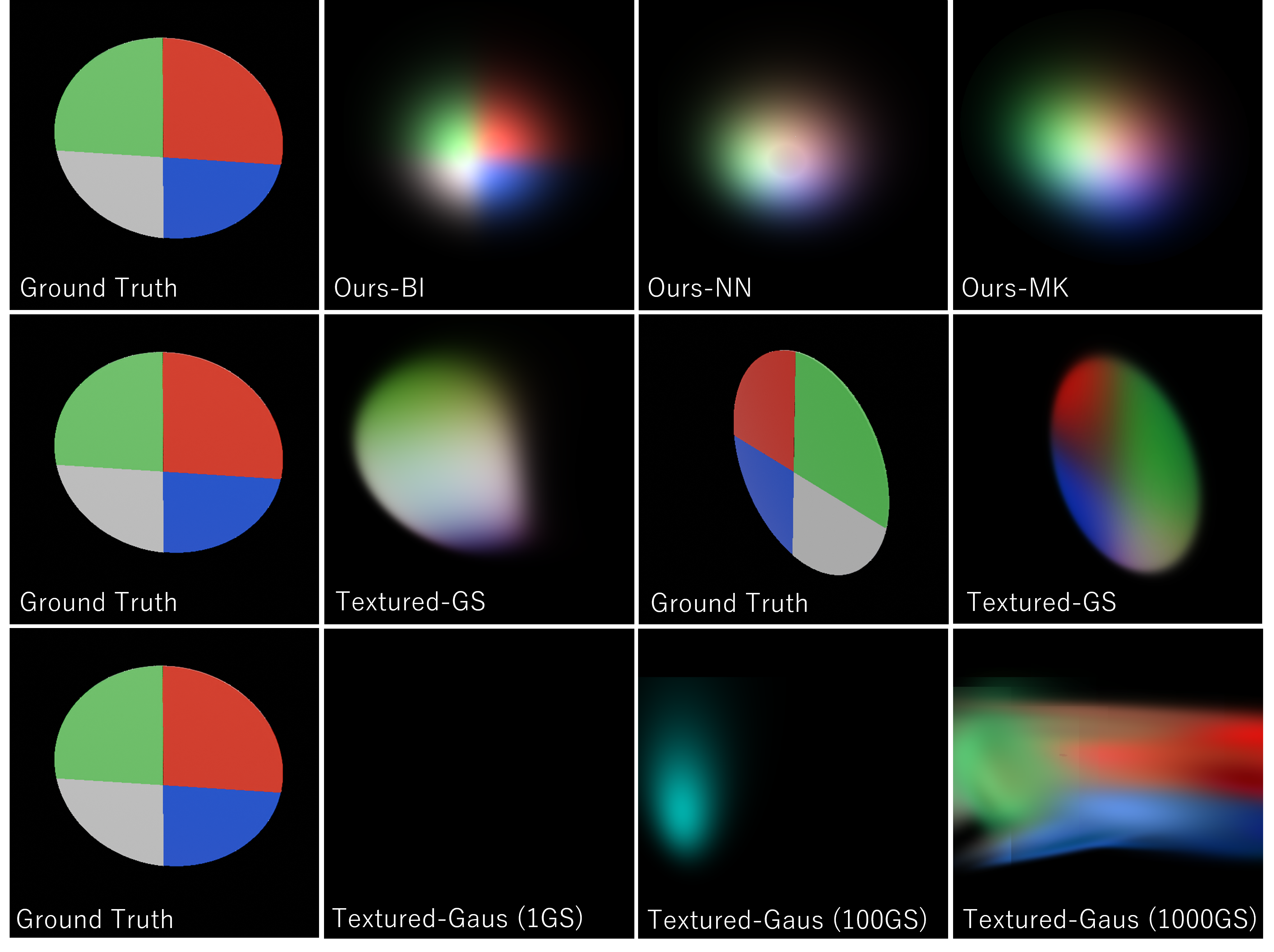}
    % \put(-1.5, 32){(a) Ours w/o Decay}
    % \put(96, 17){$\mathbf{u}$}
    % \put(77, 30){$\mathbf{v}$}
    % \put(4,  -4.0){$\lambda_s = 1.0$}
    % \put(29, -4.0){$\lambda_s = 5.0$}
    % \put(53, -4.0){$\lambda_s = 10.0$}
    % \put(84, -4.0){GT}
    \end{overpic}
    \vspace{-8mm}
    \caption{
    \XR{We illustrate the behavior of three different spatially varying functions using a single Gaussian primitive, as shown in the top row. The bilinear interpolation function captures abrupt color transitions more effectively but suffers from vanishing gradients in other regions. To further evaluate expressiveness under the constraint of a single Gaussian, we tested two additional approaches specifically designed to enhance the representational power of individual primitives. The second row presents two view results of Textured-GS~\cite{huang2024textured}, and the last row shows Textured-Gaus~\cite{chao2024textured} with different  number of Gaussian primitives.
    }
    }
\label{fig:diffff}
\vspace{-4mm}
\end{figure}
\XR{The top row of }Fig.~\ref{fig:diffff} demonstrates the image fitting capabilities of three spatial variation functions on a single Gaussian. We observe that although bilinear interpolation can better fit color mutations, \XR{it tends to experience gradient vanishing near regions of sharp change (middle axis) and is effective only when the spatial pattern conforms to a four-quadrant distribution.}
This issue arises because bilinear interpolation requires coordinates within a certain range \((0, 1)\). Therefore, we need to use the sigmoid function in Eq.~\ref{eq:sigmoid} to scale the \((u, v)\) coordinates before using them. This scaling causes the gradient of a pixel to be almost zero when it falls near the center of a quadrant, making it difficult to perfectly fit the scene. \XR{Its lagging performance on the full dataset further highlights this limitation.}

While the tiny neural network and movable kernel methods cannot fit the color mutations in the original image as precisely, they provide a smooth transition of color without causing the gradient to vanish. Consequently, when multiple Gaussians are superimposed through alpha-blending, these methods exhibit stronger expressiveness.
\FF{The slight size discrepancy from the ground truth is a natural consequence of the Gaussian splatting rasterization pipeline, where standard training typically discourages overly large individual Gaussians. The single Gaussian visualization is intended to illustrate the spatial color expressiveness within each primitive, rather than exact silhouette matching.}

\XR{Under the stringent constraint of a single Gaussian primitive, we additionally evaluated two representative methods that explicitly aim to enhance the representational capacity of individual splats. The second row of Fig.~\ref{fig:diffff} shows results from Textured-GS~\cite{huang2024textured}, which can model certain degrees of spatial color variation; however, it suffers from evident color and transparency attenuation near the center, and displays pronounced multi-view inconsistency, with its color patterns rapidly deteriorating under even slight viewpoint changes. The last row presents Textured-Gaus~\cite{chao2024textured}, which fails entirely when restricted to a single Gaussian. Even when the number of primitives is increased to 100 or 1000, the reconstructions remain visibly coarse and structurally unstable, indicating that the approach struggles to capture fine-grained appearance and geometry details even with a larger primitive budget.}

It is worth noting that for the proposed movable kernels, the kernel position moves with the gradient. In some extreme cases, the kernel center may move outside the Gaussian. At this point, \name~will degenerate into 2DGS~\cite{2dgs}. We do not impose any special restrictions on the kernel position to prevent it from moving outside the Gaussian, as this rarely occurs. In Table~\ref{tab:inside}, we present the probability of the kernel remaining inside the Gaussian across several datasets we tested. It can be seen that there is almost no instance where the kernel moves outside the Gaussian with the gradient.

\begin{table}[!tp]
    \centering
    \caption{Average percentage of kernels falling inside the Gaussian across multiple datasets.}
    \resizebox{0.9\linewidth}{!}{
\begin{tabular}{l|cccc}
\toprule
  & Synthetic Blender  & Mip-NeRF360   & Tanks\&Temples   & DTU   \\ \midrule
Internal & 99.95\% & 99.60\% & 99.92\% & 99.57\% \\ \bottomrule
\end{tabular}
}
\vspace{-2mm}

    \label{tab:inside}
\end{table}

\subsection{Discussion on Anti-aliasing}
\XR{Aliasing, commonly perceived as jagged edges or flickering when the viewing resolution or scale changes, arises when the sampling rate is insufficient to capture the high-frequency details of a scene. Anti-aliasing techniques therefore aim to suppress these artifacts by applying prefiltering or multi-scale integration before rasterization. Among recent Gaussian-based rendering methods, Mip-Splatting~\cite{yu2024mip} addresses this issue explicitly by introducing a scale-aware formulation that filters each Gaussian according to its projected footprint, effectively preventing aliasing during zoom-in/out rendering. 

\begin{table}[!tp]
    \centering
    \caption{\XR{PSNR Comparison of anti-aliasing robustness on the Blender~\cite{mildenhall2021nerf} dataset. Our method achieves superior quality to both 2DGS and 3DGS despite not being explicitly designed for anti-aliasing.}
}
    \resizebox{0.9\linewidth}{!}{
\begin{tabular}{l|ccccc}
\toprule
              & Full Res. & 1/2 Res. & 1/4 Res. & 1/8 Res. & Avg.  \\ \midrule
NeRF~\cite{mildenhall2021nerf}          & 31.48     & 32.43    & 30.29    & 26.70    & 30.23 \\
MipNeRF~\cite{barron2021mip}       & 33.08     & 33.31    & 30.91    & 27.97    & 31.31 \\
TensoRF~\cite{chen2022tensorf}       & 32.53     & 32.91    & 30.01    & 26.45    & 30.48 \\
Instant-NGP~\cite{muller2022instant}   & 33.09     & 33.00    & 29.84    & 26.33    & 30.57 \\
Tri-MipRF~\cite{hu2023tri}     & 32.89     & 32.84    & 28.29    & 23.87    & 29.47 \\ \midrule
2DGS~\cite{2dgs}          & 32.87     & 26.75    & 20.77    & 17.31    & 24.43 \\
3DGS~\cite{3dgs}          & 33.33     & 26.95    & 21.38    & 17.69    & 24.84 \\
Ours-MK          & \textbf{34.10} & \textbf{29.04} & \textbf{22.06} & \textbf{18.24} & \textbf{25.86} \\ \midrule
Mip-Splatting~\cite{yu2024mip} & 33.36     & 34.00    & 31.85    & 28.67    & 31.97 \\ \bottomrule
\end{tabular}
}
\vspace{-2mm}
    \label{tab:aa_blender}
\end{table}

Although our method is not specifically designed for anti-aliasing, we still evaluate its performance under resolution scaling to provide a fair comparison. 
Table~\ref{tab:aa_blender} reports the PSNR results on the Blender~\cite{mildenhall2021nerf} dataset at different rendering resolutions. 
Despite the lack of dedicated anti-aliasing mechanisms, our approach consistently outperforms both 2DGS~\cite{2dgs} and 3DGS~\cite{3dgs}.
3DGS generally achieves higher rendering quality than 2DGS owing to its volumetric representation, yet our method surpasses 3DGS in all tested resolution settings, demonstrating the robustness and expressiveness of spatially varying splats. 
As expected, Mip-Splatting~\cite{yu2024mip} achieves the best alias-free performance across scales due to its explicit multi-scale filtering. 
We consider integrating such scale-adaptive filtering strategies with our spatially varying splat functions as an important direction for future work, potentially combining the strengths of both approaches.}

\begin{table}[!tp]
    \centering
    \caption{
        \XR{Training cost (in seconds), rendering performance (in FPS) of our method on the Synthetic Blender~\cite{mildenhall2021nerf} dataset. 
        The right column group shows our reconstruction quality when the total training time is restricted to match that of 2DGS.}
    }
    \resizebox{\linewidth}{!}{
    \begin{tabular}{l|cc|ccc}
        \toprule
        Method & Training Time (s)$\downarrow$ & FPS$\uparrow$ & PSNR$\uparrow$ & SSIM$\uparrow$ & LPIPS$\downarrow$ \\
        \midrule
        2DGS   & 635.25 & 210.73 & 32.87 & 0.965 & 0.038 \\
        Ours-MK & 1083.13 & 133.25 & 33.77 & 0.969 & 0.031 \\
        \bottomrule
    \end{tabular}
    }
    \vspace{-2mm}
    \label{tab:time}
\end{table}

\subsection{Limitations and Future Work}
\textbf{Timing.}
\FF{
One limitation of our method is the running time. Since we introduce spatially varying function evaluations in both the forward rendering and back propagation stages, the current training and rendering speeds are slightly slower than 2DGS~\cite{2dgs}. The additional overhead mainly comes from the higher per Gaussian computation cost, as each Gaussian requires evaluating four sub kernels for weight computation, color blending, and their corresponding gradients. Table~\ref{tab:time} reports the time cost comparison with 2DGS~\cite{2dgs} on the Synthetic Blender~\cite{mildenhall2021nerf} dataset.

Nevertheless, our rendering speed still comfortably satisfies real time requirements, remaining above 30 FPS for interactive applications. Moreover, the stronger expressiveness of each Gaussian allows our method to achieve comparable or better reconstruction quality with fewer primitives. In scenes where vanilla 2DGS requires a dense set of Gaussians to model spatially varying appearance, our method can represent the same details with fewer primitives, which partially compensates for the increased per Gaussian overhead. Therefore, this trade off between computational efficiency and reconstruction fidelity is particularly attractive for detail intensive applications. Our approach is also compatible with recent acceleration frameworks, since the improved per primitive expressiveness can complement existing optimization strategies without requiring fundamental architectural changes.
\XR{We also conduct an ablation under a limited training time setting. By constraining the number of training iterations, our method can be trained within the same time budget as 2DGS~\cite{2dgs}. The results in the right half of Table~\ref{tab:time} show that, even when the training time is reduced by one third, our approach still outperforms 2DGS.}
}

\textbf{More Spatially Varying Functions.}
Another limitation is that we have not fully explored the potential of spatially varying functions. The design of spatially varying functions can be further studied. Additionally, implementing spatially varying functions on Gaussian ellipsoids~\cite{3dgs} is also an interesting research direction.

% \textbf{Extension to 3DGS and its Variants.}
% Our method builds upon 2DGS, where each primitive constitutes a plane geometry that inherently simplifies the determination of intersection coordinates for the current pixel within the Gaussian ellipsoid coordinate system. In contrast, the analogous process for 3DGS necessitates meticulous architectural design to achieve optimal performance.

\textbf{Extension to 3DGS and its Variants.}
\FF{
Our method builds upon 2DGS, where each primitive is represented as a planar geometry, which naturally simplifies the computation of local coordinates in the Gaussian coordinate system. By contrast, extending the same idea to 3DGS requires more careful design. 
As an exciting direction for future work, we will try to extend the proposed spatially varying formulation to 3DGS and related volumetric variants. At the formulation level, the current four sub kernels can be generalized to volumetric counterparts, while the core idea of learnable spatially varying sub primitives remains applicable beyond 2DGS. This extension could further improve rendering fidelity while preserving compact and explicit representations.
}
% Our method builds upon 2DGS, where each primitive constitutes a plane geometry that inherently simplifies the determination of intersection coordinates for the current pixel within the Gaussian ellipsoid coordinate system. In contrast, the analogous process for 3DGS necessitates meticulous architectural design to achieve optimal performance. 
% \XR{As an exciting direction for future work, we will try to extend the proposed spatially varying formulation to 3DGS and its volumetric variants, which could further improve rendering fidelity while maintaining compact and explicit representations.}

% For a simple implementation, we use the distance from the pixel to the projection point of the 3DGS primitive and map back to the distance on the Gaussian ellipsoid coordinate system, referring to the method of calculating the distance $\mathbf{x}$ (Eq. 4 in the original 3DGS paper) of the Gaussian function to the Gaussian center point.
% Implementing four adjustable kernels with this adaptation yields measurable improvements, elevating PSNR in floral scenes from \textit{21.52 dB} to \textit{21.61 dB}.\XR{Do we have to mention this number?} These preliminary results, obtained using a legacy 3DGS codebase, demonstrate broader applicability. By implementing the state-of-the-art 3DGS methods, the performance is expected to be improved due to the new training method and advanced sparse parameterization techniques.

\section{Conclusion}

In this paper, we introduce a new method called \name~that utilizes spatially varying colors and opacity in a single Gaussian primitive to enhance its representation ability. We propose three different spatially varying functions defined on Gaussian primitives, each of which outperforms the baseline 2DGS~\cite{2dgs}. 
\XR{Equipped with movable kernels, \name~outperforms all state-of-the-art methods on the Blender~\cite{mildenhall2021nerf} dataset, particularly excelling at representing complex textures over geometrically simple or flat regions.}
When using the normal consistency loss, we achieve geometry reconstruction quality comparable to 2DGS~\cite{2dgs} and outperform 2DGS~\cite{2dgs} when the number of Gaussians is limited, while significantly improving image rendering quality. 
Although our training and rendering speeds are slower than baseline methods due to code optimization issues, we provide a new direction for further research on Gaussian primitives and make it possible to implement more effective spatial variation functions.

% \clearpage

\bibliographystyle{unsrt}
\bibliography{main}

@String(CVPR= {IEEE Conf. Comput. Vis. Pattern Recog.})

@String(ECCV= {Eur. Conf. Comput. Vis.})

@String(TOG= {ACM Trans. Graph.})

@String(CVPR  = {CVPR})

@String(ECCV  = {ECCV})

@String(TOG   = {ACM TOG})

@article{zhou2025splat,
  title={Splat the Net: Radiance Fields with Splattable Neural Primitives},
  author={Zhou, Xilong and Nguyen, Bao-Huy and Magne, Lo{\"\i}c and Golyanik, Vladislav and Leimk{\"u}hler, Thomas and Theobalt, Christian},
  journal={The Fourteenth International Conference on Learning Representations},
  year={2025}
}

@article{3dgs,
  title={3D Gaussian Splatting for Real-Time Radiance Field Rendering.},
  author={Kerbl, Bernhard and Kopanas, Georgios and Leimk{\"u}hler, Thomas and Drettakis, George},
  journal={ACM Trans. Graph.},
  volume={42},
  number={4},
  pages={139--1},
  year={2023}
}

@inproceedings{2dgs,
  title={2d gaussian splatting for geometrically accurate radiance fields},
  author={Huang, Binbin and Yu, Zehao and Chen, Anpei and Geiger, Andreas and Gao, Shenghua},
  booktitle={ACM SIGGRAPH 2024 Conference Papers},
  pages={1--11},
  year={2024}
}

@article{mildenhall2021nerf,
  title={Nerf: Representing scenes as neural radiance fields for view synthesis},
  author={Mildenhall, Ben and Srinivasan, Pratul P and Tancik, Matthew and Barron, Jonathan T and Ramamoorthi, Ravi and Ng, Ren},
  journal={Communications of the ACM},
  volume={65},
  number={1},
  pages={99--106},
  year={2021},
  publisher={ACM New York, NY, USA}
}

@inproceedings{barron2022mipnerf360,
  title={Mip-nerf 360: Unbounded anti-aliased neural radiance fields},
  author={Barron, Jonathan T and Mildenhall, Ben and Verbin, Dor and Srinivasan, Pratul P and Hedman, Peter},
  booktitle={Proceedings of the IEEE/CVF conference on computer vision and pattern recognition},
  pages={5470--5479},
  year={2022}
}

@article{knapitsch2017tanks,
  title={Tanks and temples: Benchmarking large-scale scene reconstruction},
  author={Knapitsch, Arno and Park, Jaesik and Zhou, Qian-Yi and Koltun, Vladlen},
  journal={ACM Transactions on Graphics (ToG)},
  volume={36},
  number={4},
  pages={1--13},
  year={2017},
  publisher={ACM New York, NY, USA}
}

@inproceedings{jensen2014DTU,
  title={Large scale multi-view stereopsis evaluation},
  author={Jensen, Rasmus and Dahl, Anders and Vogiatzis, George and Tola, Engin and Aan{\ae}s, Henrik},
  booktitle={Proceedings of the IEEE conference on computer vision and pattern recognition},
  pages={406--413},
  year={2014}
}

@inproceedings{zwicker2001ewa,
  title={EWA volume splatting},
  author={Zwicker, Matthias and Pfister, Hanspeter and Van Baar, Jeroen and Gross, Markus},
  booktitle={Proceedings Visualization, 2001. VIS'01.},
  pages={29--538},
  year={2001},
  organization={IEEE}
}

@article{yariv2021volsdf,
  title={Volume rendering of neural implicit surfaces},
  author={Yariv, Lior and Gu, Jiatao and Kasten, Yoni and Lipman, Yaron},
  journal={Advances in Neural Information Processing Systems},
  volume={34},
  pages={4805--4815},
  year={2021}
}

@inproceedings{chen2022tensorf,
  title={Tensorf: Tensorial radiance fields},
  author={Chen, Anpei and Xu, Zexiang and Geiger, Andreas and Yu, Jingyi and Su, Hao},
  booktitle={European conference on computer vision},
  pages={333--350},
  year={2022},
  organization={Springer}
}

@article{muller2022instant,
  title={Instant neural graphics primitives with a multiresolution hash encoding},
  author={M{\"u}ller, Thomas and Evans, Alex and Schied, Christoph and Keller, Alexander},
  journal={ACM transactions on graphics (TOG)},
  volume={41},
  number={4},
  pages={1--15},
  year={2022},
  publisher={ACM New York, NY, USA}
}

@inproceedings{guedon2024sugar,
  title={Sugar: Surface-aligned gaussian splatting for efficient 3d mesh reconstruction and high-quality mesh rendering},
  author={Gu{\'e}don, Antoine and Lepetit, Vincent},
  booktitle={Proceedings of the IEEE/CVF Conference on Computer Vision and Pattern Recognition},
  pages={5354--5363},
  year={2024}
}

@inproceedings{kazhdan2006poisson,
  title={Poisson surface reconstruction},
  author={Kazhdan, Michael and Bolitho, Matthew and Hoppe, Hugues},
  booktitle={Proceedings of the fourth Eurographics symposium on Geometry processing},
  volume={7},
  number={4},
  year={2006}
}

@article{xu2023globally,
  title={Globally consistent normal orientation for point clouds by regularizing the winding-number field},
  author={Xu, Rui and Dou, Zhiyang and Wang, Ningna and Xin, Shiqing and Chen, Shuangmin and Jiang, Mingyan and Guo, Xiaohu and Wang, Wenping and Tu, Changhe},
  journal={ACM Transactions on Graphics (TOG)},
  volume={42},
  number={4},
  pages={1--15},
  year={2023},
  publisher={ACM New York, NY, USA}
}

@article{xu2022rfeps,
  title={Rfeps: Reconstructing feature-line equipped polygonal surface},
  author={Xu, Rui and Wang, Zixiong and Dou, Zhiyang and Zong, Chen and Xin, Shiqing and Jiang, Mingyan and Ju, Tao and Tu, Changhe},
  journal={ACM Transactions on Graphics (TOG)},
  volume={41},
  number={6},
  pages={1--15},
  year={2022},
  publisher={ACM New York, NY, USA}
}

@article{lin2022surface,
  title={Surface reconstruction from point clouds without normals by parametrizing the gauss formula},
  author={Lin, Siyou and Xiao, Dong and Shi, Zuoqiang and Wang, Bin},
  journal={ACM Transactions on Graphics},
  volume={42},
  number={2},
  pages={1--19},
  year={2022},
  publisher={ACM New York, NY}
}

@inproceedings{nan2017polyfit,
  title={Polyfit: Polygonal surface reconstruction from point clouds},
  author={Nan, Liangliang and Wonka, Peter},
  booktitle={Proceedings of the IEEE International Conference on Computer Vision},
  pages={2353--2361},
  year={2017}
}

@inproceedings{seitz2006comparison,
  title={A comparison and evaluation of multi-view stereo reconstruction algorithms},
  author={Seitz, Steven M and Curless, Brian and Diebel, James and Scharstein, Daniel and Szeliski, Richard},
  booktitle={2006 IEEE computer society conference on computer vision and pattern recognition (CVPR'06)},
  volume={1},
  pages={519--528},
  year={2006},
  organization={IEEE}
}

@inproceedings{schonberger2016structure,
  title={Structure-from-motion revisited},
  author={Schonberger, Johannes L and Frahm, Jan-Michael},
  booktitle={Proceedings of the IEEE conference on computer vision and pattern recognition},
  pages={4104--4113},
  year={2016}
}

@inproceedings{lu2024scaffold,
  title={Scaffold-gs: Structured 3d gaussians for view-adaptive rendering},
  author={Lu, Tao and Yu, Mulin and Xu, Linning and Xiangli, Yuanbo and Wang, Limin and Lin, Dahua and Dai, Bo},
  booktitle={Proceedings of the IEEE/CVF Conference on Computer Vision and Pattern Recognition},
  pages={20654--20664},
  year={2024}
}

@article{kasymov2024neggs,
  title={NegGS: Negative Gaussian Splatting},
  author={Kasymov, Artur and Czekaj, Bartosz and Mazur, Marcin and Spurek, Przemys{\l}aw},
  journal={arXiv preprint arXiv:2405.18163},
  year={2024}
}

@article{li20243d,
  title={3D-HGS: 3D Half-Gaussian Splatting},
  author={Li, Haolin and Liu, Jinyang and Sznaier, Mario and Camps, Octavia},
  journal={arXiv preprint arXiv:2406.02720},
  year={2024}
}

@inproceedings{yu2024mip,
  title={Mip-splatting: Alias-free 3d gaussian splatting},
  author={Yu, Zehao and Chen, Anpei and Huang, Binbin and Sattler, Torsten and Geiger, Andreas},
  booktitle={Proceedings of the IEEE/CVF Conference on Computer Vision and Pattern Recognition},
  pages={19447--19456},
  year={2024}
}

@inproceedings{hamdi2024ges,
  title={Ges: Generalized exponential splatting for efficient radiance field rendering},
  author={Hamdi, Abdullah and Melas-Kyriazi, Luke and Mai, Jinjie and Qian, Guocheng and Liu, Ruoshi and Vondrick, Carl and Ghanem, Bernard and Vedaldi, Andrea},
  booktitle={Proceedings of the IEEE/CVF Conference on Computer Vision and Pattern Recognition},
  pages={19812--19822},
  year={2024}
}

@inproceedings{charatan2024pixelsplat,
  title={pixelsplat: 3d gaussian splats from image pairs for scalable generalizable 3d reconstruction},
  author={Charatan, David and Li, Sizhe Lester and Tagliasacchi, Andrea and Sitzmann, Vincent},
  booktitle={Proceedings of the IEEE/CVF Conference on Computer Vision and Pattern Recognition},
  pages={19457--19467},
  year={2024}
}

@inproceedings{huang2023nerf,
  title={NeRF-texture: Texture synthesis with neural radiance fields},
  author={Huang, Yi-Hua and Cao, Yan-Pei and Lai, Yu-Kun and Shan, Ying and Gao, Lin},
  booktitle={ACM SIGGRAPH 2023 Conference Proceedings},
  pages={1--10},
  year={2023}
}

@inproceedings{Dai2024GaussianSurfels,
  author    = {Dai, Pinxuan and Xu, Jiamin and Xie, Wenxiang and Liu, Xinguo and Wang, Huamin and Xu, Weiwei},
  title     = {High-quality Surface Reconstruction using Gaussian Surfels},
  publisher = {Association for Computing Machinery},
  booktitle = {SIGGRAPH 2024 Conference Papers},
  year      = {2024},
  doi       = {10.1145/3641519.3657441}
}

@article{wang2021neus,
  title={NeuS: Learning Neural Implicit Surfaces by Volume Rendering for Multi-view Reconstruction},
  author={Wang, Peng and Liu, Lingjie and Liu, Yuan and Theobalt, Christian and Komura, Taku and Wang, Wenping},
  journal={arXiv preprint arXiv:2106.10689},
  year={2021}
}

@online{openMVS,
    title={OpenMVS: Open Multi-View Stereo reconstruction library},
    author = "OpenMVS",
    url  = "https://cdcseacave.github.io/",
}

@inproceedings{schoenberger2016colmap,
    author={Sch\"{o}nberger, Johannes Lutz and Zheng, Enliang and Pollefeys, Marc and Frahm, Jan-Michael},
    title={Pixelwise View Selection for Unstructured Multi-View Stereo},
    booktitle={European Conference on Computer Vision (ECCV)},
    year={2016},
}

@article{furukawa2010accurate,
  author={Furukawa, Yasutaka and Ponce, Jean},
  journal={IEEE Transactions on Pattern Analysis and Machine Intelligence}, 
  title={Accurate, Dense, and Robust Multiview Stereopsis}, 
  year={2010},
  volume={32},
  number={8},
  pages={1362-1376},
  doi={10.1109/TPAMI.2009.161}
}

@inproceedings{yao2018mvsnet,
  title={MVSNet: Depth inference for unstructured multi-view stereo},
  author={Yao, Yao and Luo, Zixin and Li, Shiwei and Fang, Tian and Quan, Long},
  booktitle={Proceedings of the European conference on computer vision (ECCV)},
  pages={767--783},
  year={2018}
}

@inproceedings{yao2019recurrent,
  title={Recurrent mvsnet for high-resolution multi-view stereo depth inference},
  author={Yao, Yao and Luo, Zixin and Li, Shiwei and Shen, Tianwei and Fang, Tian and Quan, Long},
  booktitle={Proceedings of the IEEE/CVF conference on computer vision and pattern recognition},
  pages={5525--5534},
  year={2019}
}

@inproceedings{luo2019pmvsnet,
  title={P-mvsnet: Learning patch-wise matching confidence aggregation for multi-view stereo},
  author={Luo, Keyang and Guan, Tao and Ju, Lili and Huang, Haipeng and Luo, Yawei},
  booktitle={Proceedings of the IEEE/CVF International Conference on Computer Vision},
  pages={10452--10461},
  year={2019}
}

@inproceedings{yu2020fast,
  title={Fast-MVSNet: Sparse-to-dense multi-view stereo with learned propagation and gauss-newton refinement},
  author={Yu, Zehao and Gao, Shenghua},
  booktitle={Proceedings of the IEEE/CVF conference on computer vision and pattern recognition},
  pages={1949--1958},
  year={2020}
}

@article{zhang2023vis,
  title={Vis-MVSNet: Visibility-aware multi-view stereo network},
  author={Zhang, Jingyang and Li, Shiwei and Luo, Zixin and Fang, Tian and Yao, Yao},
  journal={International Journal of Computer Vision},
  volume={131},
  number={1},
  pages={199--214},
  year={2023},
  publisher={Springer}
}

@inproceedings{hu2023tri,
  title={Tri-miprf: Tri-mip representation for efficient anti-aliasing neural radiance fields},
  author={Hu, Wenbo and Wang, Yuling and Ma, Lin and Yang, Bangbang and Gao, Lin and Liu, Xiao and Ma, Yuewen},
  booktitle={Proceedings of the IEEE/CVF International Conference on Computer Vision},
  pages={19774--19783},
  year={2023}
}

@inproceedings{barron2021mip,
  title={Mip-nerf: A multiscale representation for anti-aliasing neural radiance fields},
  author={Barron, Jonathan T and Mildenhall, Ben and Tancik, Matthew and Hedman, Peter and Martin-Brualla, Ricardo and Srinivasan, Pratul P},
  booktitle={Proceedings of the IEEE/CVF International Conference on Computer Vision},
  pages={5855--5864},
  year={2021}
}

@inproceedings{barron2022mip,
  title={Mip-nerf 360: Unbounded anti-aliased neural radiance fields},
  author={Barron, Jonathan T and Mildenhall, Ben and Verbin, Dor and Srinivasan, Pratul P and Hedman, Peter},
  booktitle={Proceedings of the IEEE/CVF Conference on Computer Vision and Pattern Recognition},
  pages={5470--5479},
  year={2022}
}

@inproceedings{barron2023zip,
  title={Zip-nerf: Anti-aliased grid-based neural radiance fields},
  author={Barron, Jonathan T and Mildenhall, Ben and Verbin, Dor and Srinivasan, Pratul P and Hedman, Peter},
  booktitle={Proceedings of the IEEE/CVF International Conference on Computer Vision},
  pages={19697--19705},
  year={2023}
}

@inproceedings{fridovich2022plenoxels,
  title={Plenoxels: Radiance fields without neural networks},
  author={Fridovich-Keil, Sara and Yu, Alex and Tancik, Matthew and Chen, Qinhong and Recht, Benjamin and Kanazawa, Angjoo},
  booktitle={Proceedings of the IEEE/CVF Conference on Computer Vision and Pattern Recognition},
  pages={5501--5510},
  year={2022}
}

@article{brunet2011mathematical,
  title={On the mathematical properties of the structural similarity index},
  author={Brunet, Dominique and Vrscay, Edward R and Wang, Zhou},
  journal={IEEE Transactions on Image Processing},
  volume={21},
  number={4},
  pages={1488--1499},
  year={2011},
  publisher={IEEE}
}

@inproceedings{zhang2018perceptual,
  title={The Unreasonable Effectiveness of Deep Features as a Perceptual Metric},
  author={Zhang, Richard and Isola, Phillip and Efros, Alexei A and Shechtman, Eli and Wang, Oliver},
  booktitle={CVPR},
  year={2018}
}

@inproceedings{xu2024texture,
  title={Texture-gs: Disentangling the geometry and texture for 3d gaussian splatting editing},
  author={Xu, Tian-Xing and Hu, Wenbo and Lai, Yu-Kun and Shan, Ying and Zhang, Song-Hai},
  booktitle={European Conference on Computer Vision},
  pages={37--53},
  year={2024},
  organization={Springer}
}

@article{huang2024textured,
  title={Textured-GS: Gaussian Splatting with Spatially Defined Color and Opacity},
  author={Huang, Zhentao and Gong, Minglun},
  journal={arXiv preprint arXiv:2407.09733},
  year={2024}
}

@article{chao2024textured,
  title={Textured Gaussians for Enhanced 3D Scene Appearance Modeling},
  author={Chao, Brian and Tseng, Hung-Yu and Porzi, Lorenzo and Gao, Chen and Li, Tuotuo and Li, Qinbo and Saraf, Ayush and Huang, Jia-Bin and Kopf, Johannes and Wetzstein, Gordon and others},
  journal={arXiv preprint arXiv:2411.18625},
  year={2024}
}

@inproceedings{lee2024compact,
  title={Compact 3d gaussian representation for radiance field},
  author={Lee, Joo Chan and Rho, Daniel and Sun, Xiangyu and Ko, Jong Hwan and Park, Eunbyung},
  booktitle={Proceedings of the IEEE/CVF Conference on Computer Vision and Pattern Recognition},
  pages={21719--21728},
  year={2024}
}

@article{fang2024mini,
  title={Mini-Splatting2: Building 360 Scenes within Minutes via Aggressive Gaussian Densification},
  author={Fang, Guangchi and Wang, Bing},
  journal={arXiv preprint arXiv:2411.12788},
  year={2024}
}

@article{fan2025lightgaussian,
  title={Lightgaussian: Unbounded 3d gaussian compression with 15x reduction and 200+ fps},
  author={Fan, Zhiwen and Wang, Kevin and Wen, Kairun and Zhu, Zehao and Xu, Dejia and Wang, Zhangyang and others},
  journal={Advances in neural information processing systems},
  volume={37},
  pages={140138--140158},
  year={2025}
}

@inproceedings{mallick2024taming,
  title={Taming 3dgs: High-quality radiance fields with limited resources},
  author={Mallick, Saswat Subhajyoti and Goel, Rahul and Kerbl, Bernhard and Steinberger, Markus and Carrasco, Francisco Vicente and De La Torre, Fernando},
  booktitle={SIGGRAPH Asia 2024 Conference Papers},
  pages={1--11},
  year={2024}
}

@article{chen2024pgsr,
  title={Pgsr: Planar-based gaussian splatting for efficient and high-fidelity surface reconstruction},
  author={Chen, Danpeng and Li, Hai and Ye, Weicai and Wang, Yifan and Xie, Weijian and Zhai, Shangjin and Wang, Nan and Liu, Haomin and Bao, Hujun and Zhang, Guofeng},
  journal={IEEE Transactions on Visualization and Computer Graphics},
  year={2024},
  publisher={IEEE}
}

@article{kheradmand20243d,
  title={3d gaussian splatting as markov chain monte carlo},
  author={Kheradmand, Shakiba and Rebain, Daniel and Sharma, Gopal and Sun, Weiwei and Tseng, Yang-Che and Isack, Hossam and Kar, Abhishek and Tagliasacchi, Andrea and Yi, Kwang Moo},
  journal={Advances in Neural Information Processing Systems},
  volume={37},
  pages={80965--80986},
  year={2024}
}

@article{weiss2024gaussian,
  title={Gaussian billboards: Expressive 2d gaussian splatting with textures},
  author={Weiss, Sebastian and Bradley, Derek},
  journal={arXiv preprint arXiv:2412.12734},
  year={2024}
}

% \bf{If you include a photo:}
\vspace{-53pt}
\begin{IEEEbiography}[{\includegraphics[width=1in,height=1.25in,clip,keepaspectratio]{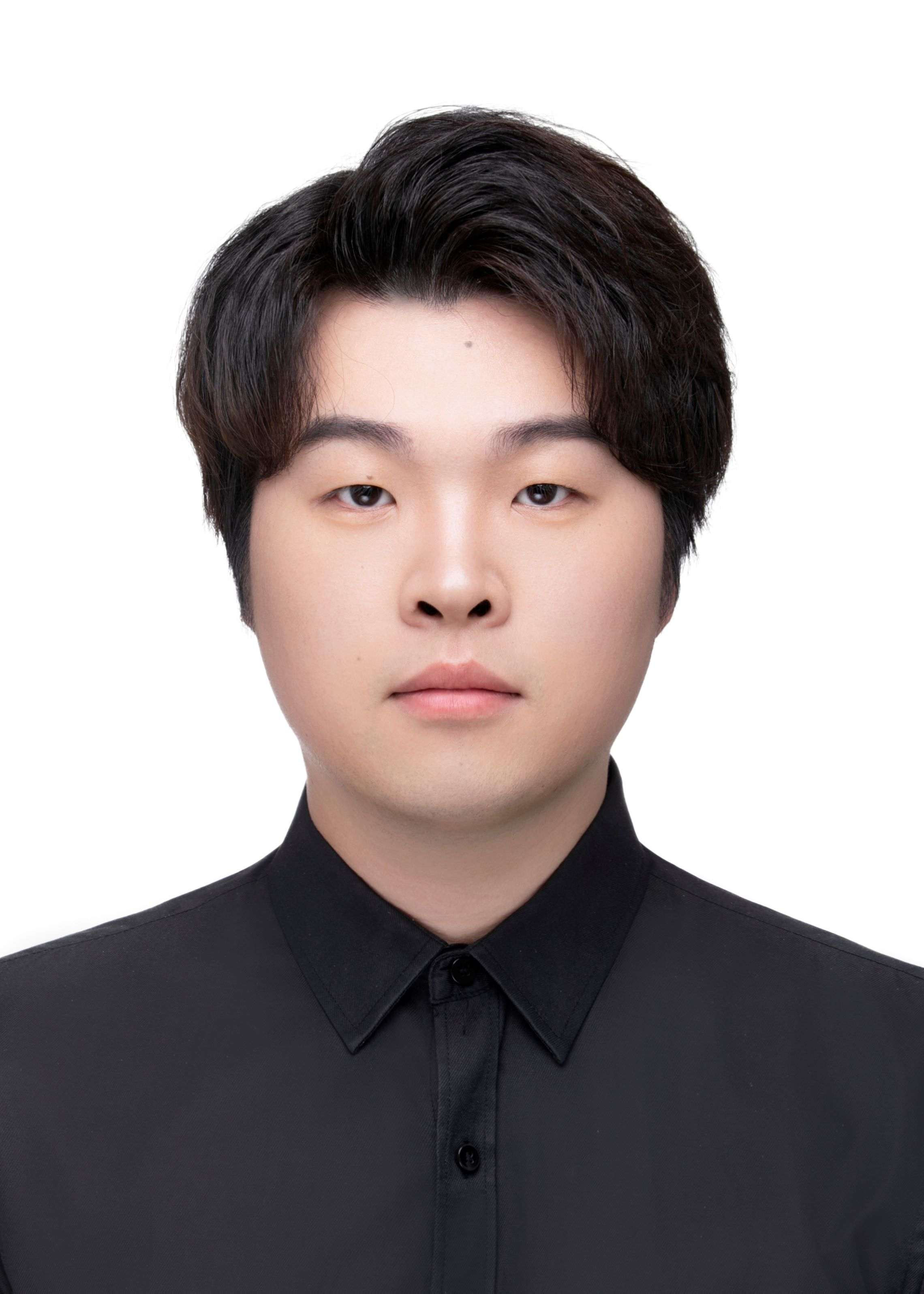}}]{Rui Xu}
is currently a Ph.D. candidate in the Department of Computer Science at The University of Hong Kong, supervised by Prof. Wenping Wang and Prof. Taku Komura. He received the M. Eng. and B. Eng. degrees from the Interdisciplinary Research Center (IRC) of Shandong University. His research interests focus on Computer Graphics, 3D Vision, Geometry Processing and Generative Models. He received the Best Paper Award at SIGGRAPH 2023.
\end{IEEEbiography}
\vspace{-33pt}
\begin{IEEEbiography}[{\includegraphics[width=1in,height=1.25in,clip,keepaspectratio]{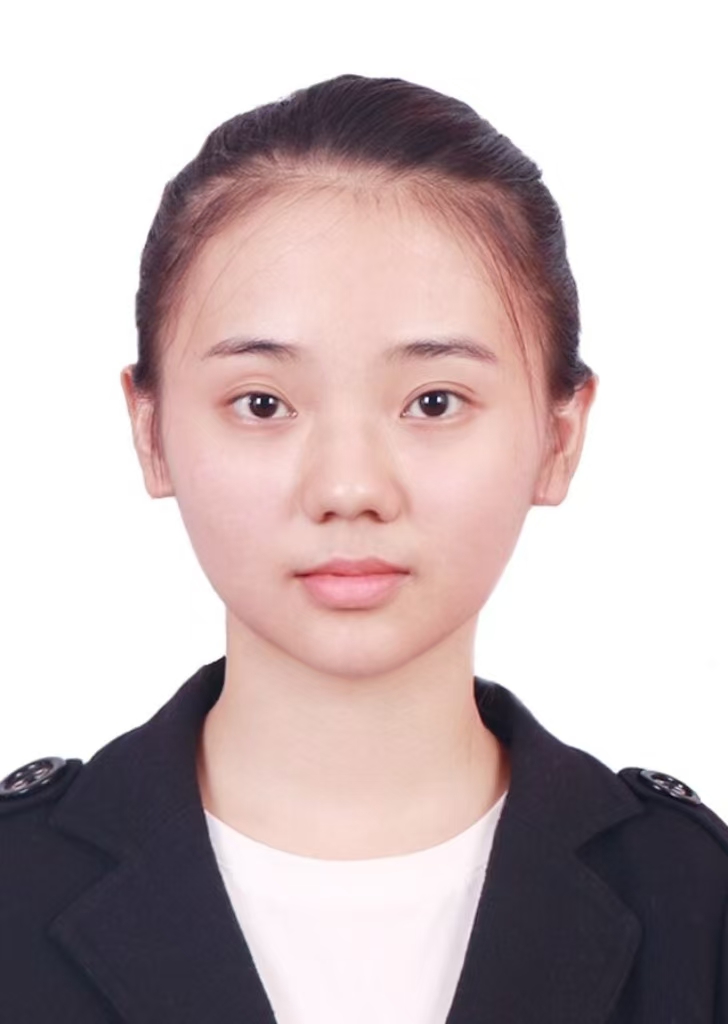}}]{Wenyue Chen} 
is currently a master’s student at Peking University, advised by Prof. Ronggang Wang. She received her B.S. degree in Artificial Intelligence from Dalian University of Technology in 2025. Her research interests focus on 3D generation.

\end{IEEEbiography}
\vspace{-33pt}
\begin{IEEEbiography}[{\includegraphics[width=1in,height=1.25in,clip,keepaspectratio]{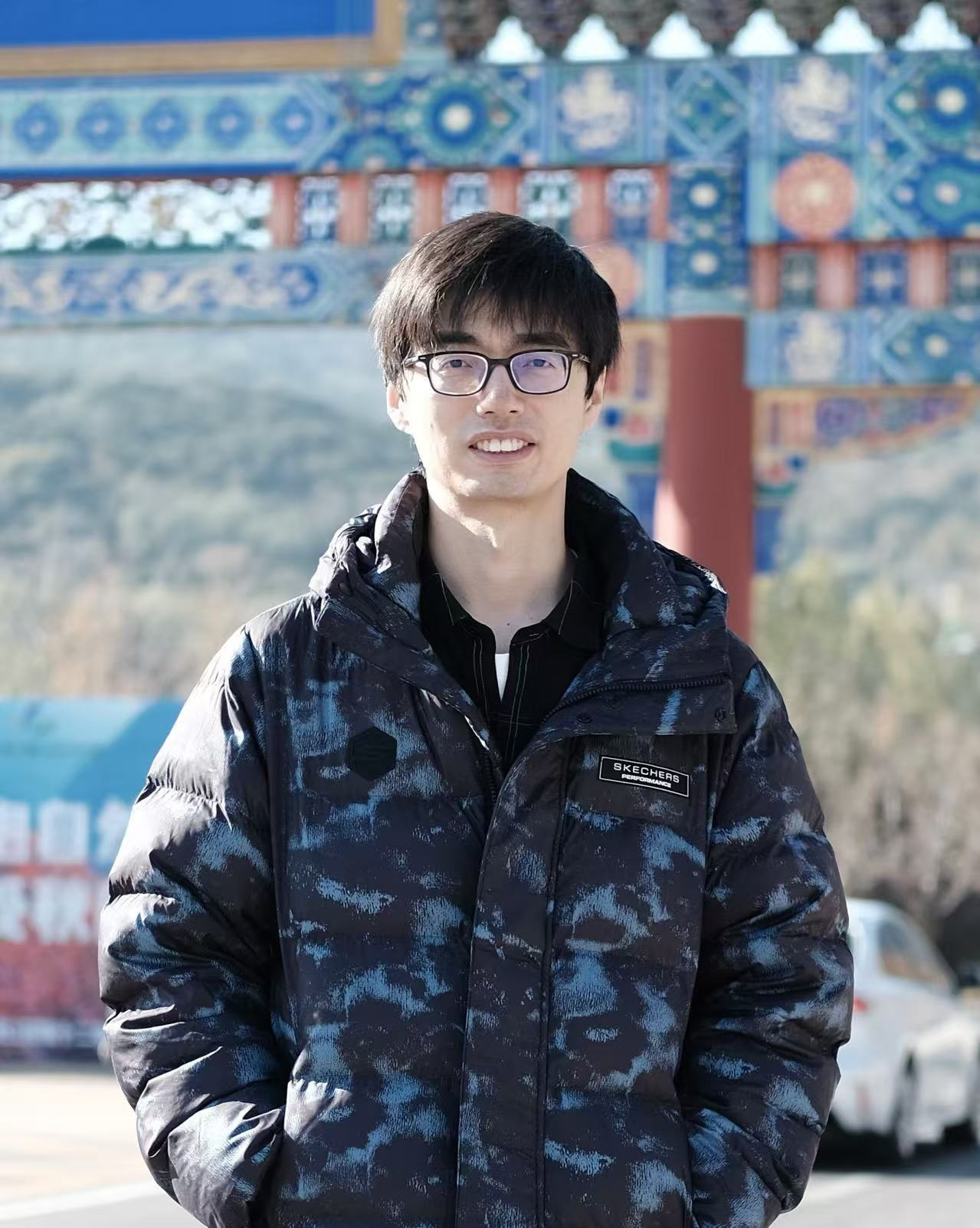}}]{Jiepeng Wang}
obtained his PhD degree at The University of Hong Kong, supervised by Prof. Wenping Wang and Prof. Taku Komura. Before that, he got his master degree at Shanghai Jiao Tong University and bachelor degree at Shandong University. He was also a research intern at MSRA working on 3D structural understanding. His research interests include image/video generation, 3D vision and computer graphics.
\end{IEEEbiography}
\vspace{-33pt}

\begin{IEEEbiography}[{\includegraphics[width=1in,height=1.25in,clip,keepaspectratio]{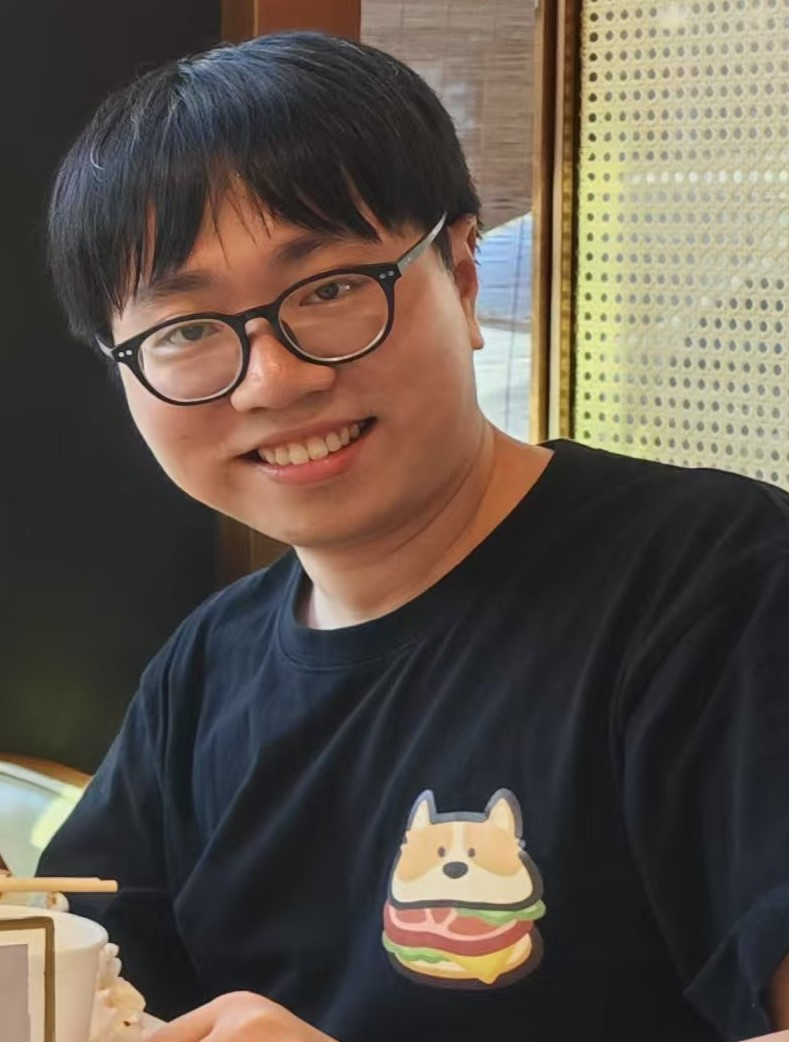}}]{Yuan Liu}
is an assistant professor at ISD HKUST. Prior to that, Yuan worked in NTU as a PostDoc researcher and obtained his PhD degree at HKU. His research mainly concentrates on 3D vision and graphics. He currently works on topics about 3D AIGC including 3D neural representations, 3D generative models, and 3D-aware video generation.
\end{IEEEbiography}
\vspace{-35pt}

\begin{IEEEbiography}[{\includegraphics[width=1in,height=1.25in,clip,keepaspectratio]{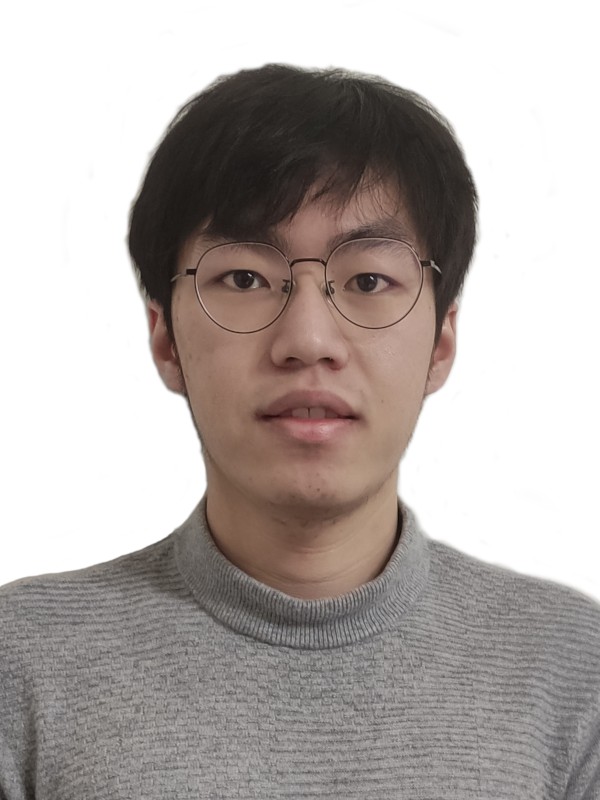}}]{Peng Wang}
obtained his PhD degree at The University of Hong Kong, supervised by Prof. Wenping Wang and Prof. Taku Komura. He was an intern at Adobe Research, working with Dr. Kai Zhang, and was a visiting student at Nanyang Technological University, advised by Prof. Ziwei Liu. His research is related to computer graphics and 3D computer vision, including 3D reconstruction, neural rendering, and 3D content creation.
\end{IEEEbiography}
\vspace{-35pt}
\begin{IEEEbiography}[{\includegraphics[width=1in,height=1.25in,clip,keepaspectratio]{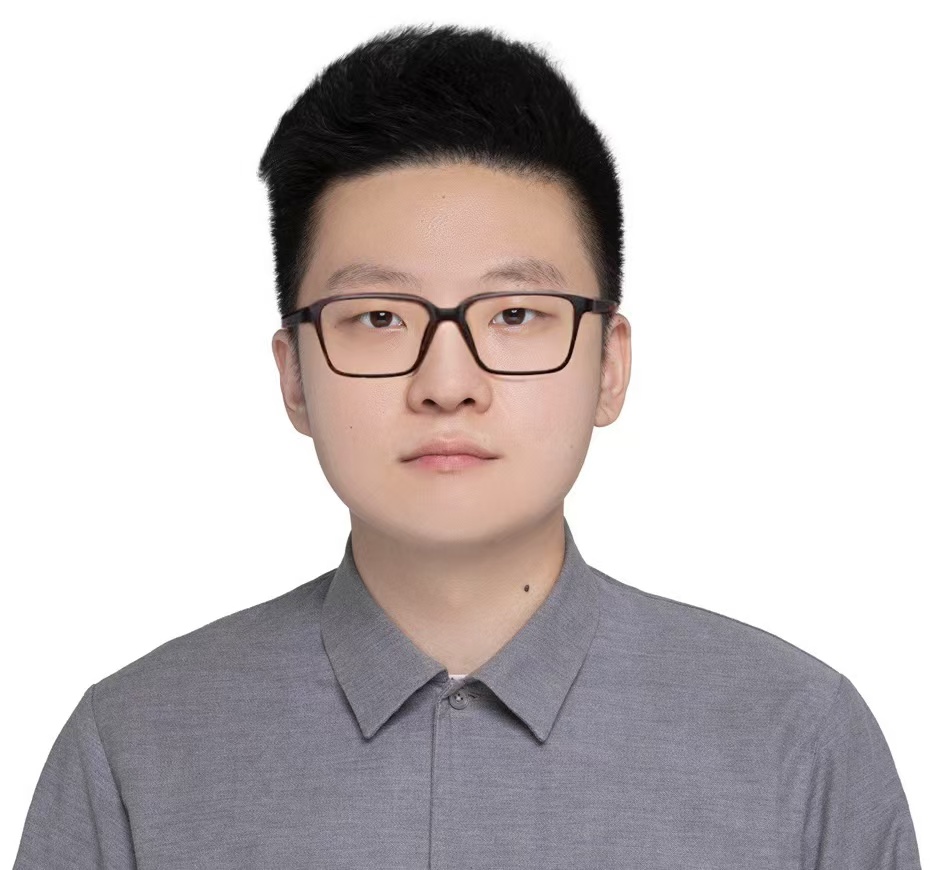}}]{Cheng Lin}
is an assistant professor at the Department of Computer Science and Engineering of Macau University of Science and Technology (MUST). Before that, he was a researcher at Tencent and miHoYo, respectively. He received his Ph.D. from The University of Hong Kong (HKU). He visited the Visual Computing Group at Technical University of Munich (TUM). His research interests include 3D vision, computer graphics, geometric modeling and shape analysis.
\end{IEEEbiography}
\vspace{-35pt}
\begin{IEEEbiography}[{\includegraphics[width=1in,height=1.25in,clip,keepaspectratio]{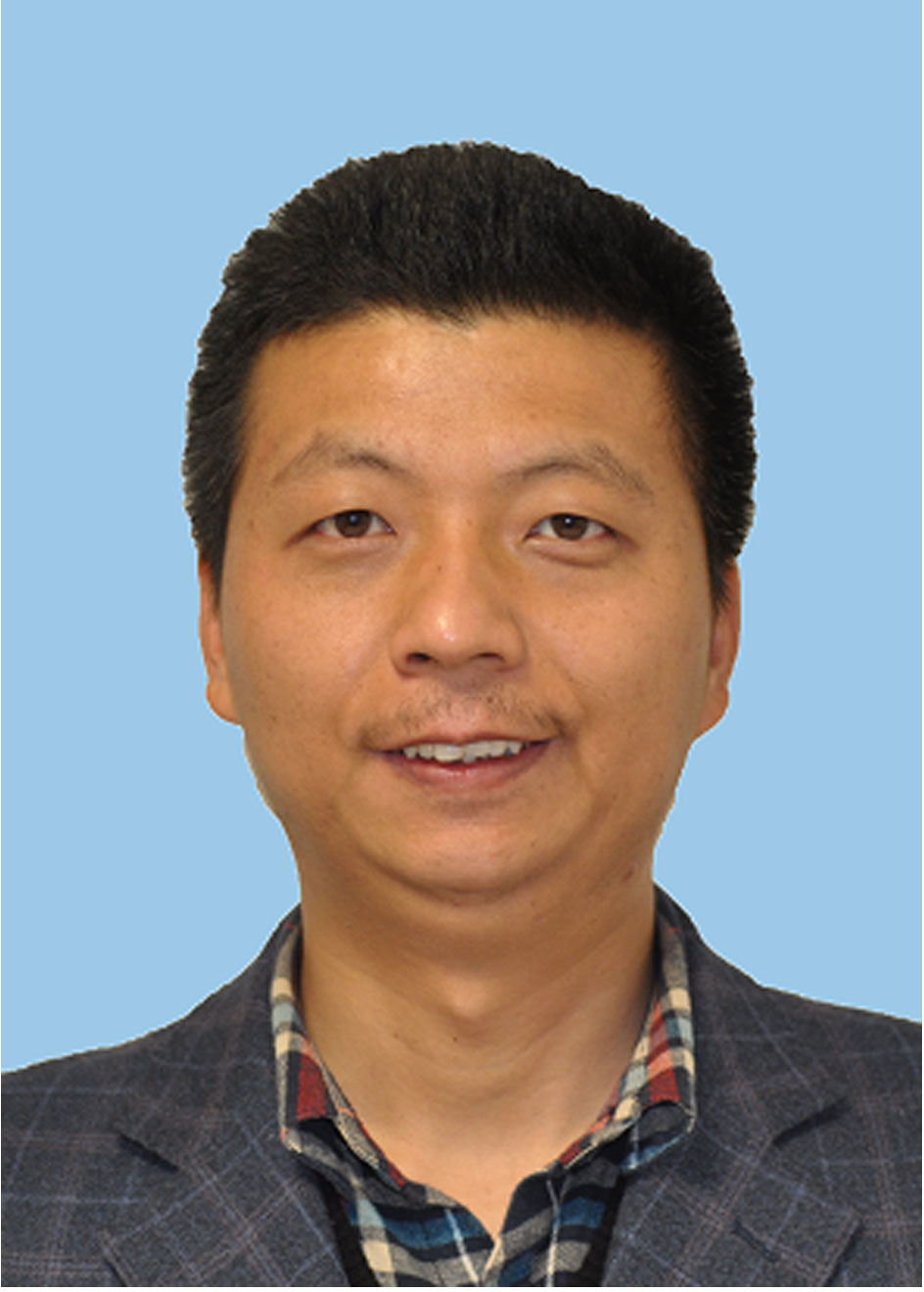}}]{Shiqing Xin} is currently a professor in the School of Computer Science at Shandong University. He obtained his Ph.D. degree from Zhejiang University in 2009. After that, he worked as a research fellow at Nangyang Technological University for three years. His research interests encompass a range of geometry processing algorithms. He has authored and co-authored over 100 papers published in renowned journals and conferences, including IEEE TVCG and ACM TOG, among others. He received the Best Paper Award at SIGGRAPH 2023.
\end{IEEEbiography}
\vspace{-34pt}

\begin{IEEEbiography}[{\includegraphics[width=1in,height=1.25in,clip,keepaspectratio]{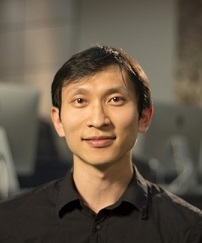}}]{Xin Li} is a Professor at the Section of Visual
Computing and Interactive Media, School of Performance, Visualization, and Fine Arts, and a joint
faculty member of Department of Computer Science
and Engineering, at Texas A\&M University. He got
his B.S. degree in Computer Science at University
of Science and Technology of China (USTC) in
2003, and Ph.D. in Computer Science from State
University of New York at Stony Brook in 2008.
His research areas are in visual computing, geometric modeling and processing, computer vision, and
computer-aided design. 
\end{IEEEbiography}
\vspace{-34pt}
\begin{IEEEbiography}[{\includegraphics[width=1in,height=1.25in,clip,keepaspectratio]{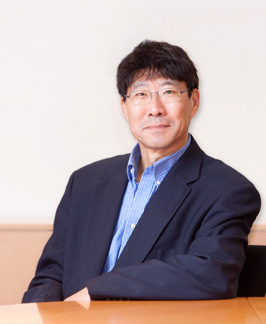}}]{Wenping Wang}
(Fellow, IEEE) received the PhD degree in computer science from the University of Alberta, in 1992. He is currently a professor of Department of Computer Science \& Engineering with Texas A\&M University. His research interests include computer graphics, computer visualization, computer vision, robotics, medical image processing, and geometric computing. He is associate editor of several premium journals, including the Computer Aided Geometric Design (CAGD), Computer Graphics Forum (CGF), IEEE Transactions on Computers, and IEEE Transactions on Visualization and
Computer Graphics, and has chaired a number of international conferences, including Pacific Graphics 2012, ACM Symposium on Physical and Solid Modeling (SPM) 2013, SIGGRAPH Asia 2013, and Geometry Submit 2019. He received the John Gregory Memorial Award and Pierre Bézier Award for his contributions in geometric modeling, and he is an ACM Fellow.
\end{IEEEbiography}
\vspace{-34pt}
\begin{IEEEbiography}[{\includegraphics[width=1in,height=1.25in,clip,keepaspectratio]{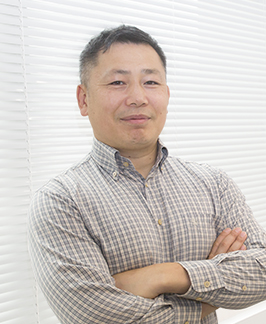}}]{Taku Komura} is a professor in the Department of Computer Science, The University of Hong Kong. Before joining The University of Hong Kong in 2020, he worked at the University of Edinburgh (2006-2020), City University of Hong Kong (2002-2006) and RIKEN (2000-2002). He received his BSc, MSc and PhD in Information Science from University of Tokyo. His research has focused on data-driven character animation, physically-based animation, crowd simulation, 3D modelling, cloth animation, anatomy-based modelling and robotics. Recently, his main research interests have been on physically-based animation and the application of machine learning techniques for animation synthesis. He received the Royal Society Industry Fellowship (2014), the Google AR/VR Research Award (2017) and the SIGGRAPH Best Paper Award (2022).  
\end{IEEEbiography}
% \vspace{11pt}

% \bf{If you will not include a photo:}\vspace{-33pt}
% \begin{IEEEbiographynophoto}{John Doe}
% Use $\backslash${\tt{begin\{IEEEbiographynophoto\}}} and the author name as the argument followed by the biography text.
% \end{IEEEbiographynophoto}

\clearpage

\vfill

\end{document}